%% file: main.tex
\documentclass[conference]{IEEEtran}
\IEEEoverridecommandlockouts
\pdfoutput=1
\usepackage{cite}
\usepackage{amsmath,amssymb,amsfonts}
\usepackage{algorithmic}
\usepackage{graphicx}
\usepackage{textcomp}
\usepackage{xcolor}
\def\BibTeX{{\rm B\kern-.05em{\sc i\kern-.025em b}\kern-.08em
    T\kern-.1667em\lower.7ex\hbox{E}\kern-.125emX}}

\input{defines}

\input{long_defines}
\usepackage{placeins}

\begin{document}

\title{Concise and interpretable multi-label rule sets}

\author{\IEEEauthorblockN{ Martino Ciaperoni}
\IEEEauthorblockA{\textit{Department of Computer Science} \\
\textit{Aalto University}\\
Espoo, Finland \\
martino.ciaperoni@aalto.fi}
\and
\IEEEauthorblockN{Han Xiao}
\IEEEauthorblockA{\textit{Department of Computer Science} \\
\textit{Aalto University}\\
Espoo, Finland \\
han.xiao@aalto.fi}
\and
\IEEEauthorblockN{Aristides Gionis}
\IEEEauthorblockA{\textit{Division of Theoretical Computer Science} \\
\textit{KTH Royal Institute of Technology}\\
Stockholm, Sweden \\
argioni@kth.se}
}

\maketitle

\input{abs.tex}

\input{intro.tex}
\input{related.tex}

\input{problem_formulation.tex}
\input{algorithm.tex}

\input{samp.tex}

\input{enhanced_sampling.tex}

\input{complexity.tex}

\input{exp.tex}

\input{conclusion.tex}
\FloatBarrier

\bibliographystyle{abbrv}
\bibliography{ref} 
\input{appendix}

\end{document}

%% file: defines.tex
\usepackage[utf8]{inputenc} %
\usepackage[T1]{fontenc}    %
\usepackage{hyperref}       %
\usepackage{url}            %
\usepackage{booktabs}       %
\usepackage{amsfonts}       %
\usepackage{nicefrac}       %
\usepackage{microtype}      %
\usepackage{amsmath}
\usepackage{amssymb}
\usepackage{mathtools}
\usepackage{graphicx}
\usepackage{url}
\usepackage{xspace}
\usepackage{float}
\usepackage{tikz}
\usetikzlibrary{calc,positioning}
\usepackage{varwidth}
\usepackage{caption}
\usepackage{enumitem}
\usepackage{pgfplots}
\pgfplotsset{compat=newest}
\usepgfplotslibrary{groupplots}
\usepgfplotslibrary{dateplot}

\makeatletter
\newcommand\footnoteref[1]{\protected@xdef\@thefnmark{\ref{#1}}\@footnotemark}
\makeatother

\usepackage[ruled, vlined, nofillcomment, linesnumbered]{algorithm2e}

\DeclareMathOperator*{\argmax}{\arg\max}

\SetKwComment{Comment}{/*}{*/}

\newtheorem{proposition}{Proposition}

\newtheorem{problem}{Problem}

\newcommand{\para}[1]{\noindent{\bf{#1}}}
\newcommand{\spara}[1]{\smallskip\noindent{\bf{#1}}}
\newcommand{\set}[1]{\left\{#1\right\}}
\newcommand{\abs}[1]{{\left|#1\right|}}

\newcommand{\real}{\ensuremath{\mathbb{R}}\xspace}

\newcommand{\pr}[1]{\ensuremath{\left(#1\right)}\xspace}
\newcommand{\spr}[1]{\left[#1\right]}

\newcommand{\np}{\ensuremath{\mathbf{NP}}\xspace}
\newcommand{\bigO}{\ensuremath{\mathcal{O}}\xspace}

\newcommand{\integer}{\ensuremath{\mathbb{Z}_{+}}\xspace}

\newcommand{\featspace}{\ensuremath{\mathcal{F}}\xspace}

\newcommand{\labelspace}{\ensuremath{\mathcal{L}}\xspace}

\newcommand{\dataset}{\ensuremath{\mathcal{D}}\xspace}
\newcommand{\ds}{\dataset}
\newcommand{\dsdef}{\ensuremath{\dataset=\set{\dr_i}_{i=1}^{\numpts}}\xspace}

\newcommand{\positiveD}{\ensuremath{D^+}\xspace}
\newcommand{\posD}{\positiveD}
\newcommand{\negativeD}{\ensuremath{D^-}\xspace}
\newcommand{\negD}{\negativeD}
\newcommand{\wgtpos}{\ensuremath{\wgt_1}}
\newcommand{\wgtneg}{\ensuremath{\wgt_2}}

\newcommand{\tuple}{\ensuremath{\mathbf{D}}\xspace}
\newcommand{\tupledef}{\ensuremath{\pr{\posD, \negD}}\xspace}

\newcommand{\positives}{\ensuremath{\mathcal{D}^+_\rtail}\xspace}
\newcommand{\negatives}{\ensuremath{\mathcal{D}^-_\rtail}\xspace}

\newcommand{\patternspace}{\ensuremath{\mathcal{S}}\xspace}
\newcommand{\samplespace}{\patternspace}
\newcommand{\reducedss}{\ensuremath{\patternspace^{-}}\xspace}

\newcommand{\dr}{\ensuremath{D}\xspace}  %
\newcommand{\numpts}{\ensuremath{n}\xspace}
\newcommand{\feats}{\ensuremath{F}\xspace}

\newcommand{\labels}{\ensuremath{L}\xspace}
\newcommand{\labelsD}{\ensuremath{\labels_{\dr}}\xspace}
\newcommand{\featsD}{\ensuremath{\feats_{\dr}}\xspace}

\newcommand{\drOne}{\ensuremath{\dr^*}\xspace}  %

\newcommand{\groundset}{\ensuremath{\mathcal{E}}\xspace}

\newcommand{\crule}{\ensuremath{R}\xspace}  %
\newcommand{\rhead}{\ensuremath{H}\xspace}

\newcommand{\rtail}{\ensuremath{T}\xspace}

\newcommand{\tailr}{\ensuremath{\rtail_{\crule}}\xspace}
\newcommand{\headr}{\ensuremath{\rhead_{\crule}}\xspace}

\newcommand{\ruledef}{\ensuremath{\crule=(\rhead \rightarrow \rtail)}\xspace}
\newcommand{\ruleset}{\ensuremath{\mathcal{\crule}}\xspace} 
\newcommand{\candidaterules}{\ensuremath{\mathcal{C}_R}\xspace} 
\newcommand{\currentrules}{\ensuremath{\mathcal{C}}\xspace} 

\newcommand{\dsX}{\ensuremath{\dataset\spr{X}\xspace}}

\newcommand{\dsR}{\ensuremath{\dataset\spr{\crule}\xspace}}
\newcommand{\dsH}{\ensuremath{\dataset\spr{\rhead}}\xspace}

\newcommand{\dsT}{\ensuremath{\dataset\spr{\rtail}}\xspace}

\newcommand{\budget}{\ensuremath{B}\xspace}

\newcommand{\quality}{\ensuremath{q}\xspace}
\newcommand{\dist}{\ensuremath{d}\xspace}
\newcommand{\disc}{\ensuremath{a}\xspace}
\newcommand{\obj}{\ensuremath{f}\xspace}

\newcommand{\area}{\ensuremath{\text{area}\!}\xspace}
\newcommand{\relarea}[1]{\ensuremath{\area\pr{#1; \ruleset}}\xspace}

\newcommand{\indicator}{\ensuremath{I\!}\xspace}
\newcommand{\Dkl}{\ensuremath{D_{\text{KL}}\!}\xspace}

\newcommand{\bindist}{\ensuremath{\text{Bin}\!}\xspace}
\newcommand{\PdsR}{\ensuremath{P_{\dsR}}\xspace}
\newcommand{\Pds}{\ensuremath{P_{\ds}}\xspace}

\newcommand{\coverage}{\ensuremath{\text{cov}\!}\xspace}
\newcommand{\cov}{\coverage}
\newcommand{\covD}{\ensuremath{\cov_{\dr}}\xspace}
\newcommand{\matches}{\ensuremath{\text{ matches }}\xspace}

\newcommand{\powerset}{\ensuremath{\mathcal{P}}\xspace}

\newcommand{\rulespace}{\ensuremath{\mathcal{U}}\xspace}

\newcommand{\wgt}{\ensuremath{w}\xspace}
\newcommand{\wgtR}[1]{\ensuremath{\wgt\!\pr{#1 ; \ruleset}}\xspace}

\newcommand{\invmap}{\ensuremath{I}\xspace}

\newcommand{\proba}[1]{\ensuremath{\text{Pr}\pr{#1}}\xspace}
\newcommand{\propodist}{\ensuremath{\overline{\wgt}}}
\newcommand{\proposalstate}{\ensuremath{\mathbf{C}_t}\xspace}

\newcommand{\covDR}{\ensuremath{\covD\pr{\ruleset}}}

\newcommand{\relcov}[1]{\ensuremath{\cov_{\dr}\!\pr{#1; \ruleset}}\xspace}

\newcommand{\relcovR}{\ensuremath{\overline{\cov}_{\dr}\!\pr{\ruleset}}\xspace}
\newcommand{\tol}{\ensuremath{\tau}\xspace}

\newcommand{\qdist}{\ensuremath{q_{a}\!}\xspace}
\newcommand{\qsupp}{\ensuremath{q_{\text{supp}}\!}\xspace}
\newcommand{\qarea}{\ensuremath{q_{\text{area}}\!}\xspace}
\newcommand{\cftp}{{\textsc{cftp}}\xspace}
\newcommand{\mcmc}{{\textsc{mcmc}}\xspace}

\newcommand{\ug}{\ensuremath{\mathcal{G}}\xspace}

\newcommand{\nodes}{\ensuremath{V}\xspace}
\newcommand{\edges}{\ensuremath{E}\xspace}
\newcommand{\ewgt}{\ensuremath{p}\xspace}

\newcommand{\ugdef}{\ensuremath{\ug=\pr{\nodes, \edges, \ewgt}}}

\newcommand{\featspool}{\ensuremath{\feats^{'}}\xspace}
\newcommand{\currH}{\ensuremath{\rhead}\xspace}
\newcommand{\discrscores}{\ensuremath{Q_{\text{disc}}}\xspace}

\newcommand{\bibtex}{\ensuremath{\textsf{\small bibtex}}\xspace}
\newcommand{\medical}{\ensuremath{\textsf{\small medical}}\xspace}
\newcommand{\enron}{\ensuremath{\textsf{\small enron}}\xspace}
\newcommand{\birds}{\ensuremath{\textsf{\small birds}}\xspace}
\newcommand{\emotions}{\ensuremath{\textsf{\small emotions}}\xspace}

\newcommand{\corel}{\ensuremath{\textsf{\small corel-5k}}\xspace}

\newcommand{\bibtexf}{\ensuremath{\textsf{\footnotesize bibtex}}\xspace}
\newcommand{\medicalf}{\ensuremath{\textsf{\footnotesize medical}}\xspace}
\newcommand{\enronf}{\ensuremath{\textsf{\footnotesize enron}}\xspace}
\newcommand{\birdsf}{\ensuremath{\textsf{\footnotesize birds}}\xspace}
\newcommand{\emotionsf}{\ensuremath{\textsf{\footnotesize emotions}}\xspace}
\newcommand{\mediamillf}{\ensuremath{\textsf{\footnotesize mediamill}}\xspace}
\newcommand{\CALf}{\ensuremath{\textsf{\footnotesize CAL500}}\xspace}
\newcommand{\corelf}{\ensuremath{\textsf{\footnotesize corel-5k}}\xspace}
\newcommand{\Yelpf}{\ensuremath{\textsf{\footnotesize Yelp}}\xspace}

\newcommand{\Boomer}{\textsc{Boomer}\xspace}
\newcommand{\seco}{\textsc{SeCo}\xspace}
\newcommand{\SVM}{\textsc{svm-br}\xspace}
\newcommand{\ouralgorithm}{\textsc{corset}\xspace}
\newcommand{\ouralg}{\ouralgorithm}
\newcommand{\samplecandalg}{\textsc{Gen\-Cand\-Rules}\xspace}

\newcommand{\ouralgorithmsurs}{\ouralgorithm-\textsc{surs}\xspace}
\newcommand{\ouralgsurs}{\ouralgorithmsurs\xspace}
\newcommand{\ouralggh}{\ouralgorithm-\textsc{gh}\xspace}
\newcommand{\ouralgorithmplus}{\ouralgorithm-\textsc{gh}\xspace}

\newcommand{\nullobj}{\ensuremath{\perp}\xspace}

%% file: long_defines.tex
\tikzset{%
  point/.style={circle, inner sep=2pt}, 
  other point/.style={fill=black, point}  
}

\makeatletter
\providecommand*{\cupdot}{%
  \mathbin{%
    \mathpalette\@cupdot{}%
  }%
}
\newcommand*{\@cupdot}[2]{%
  \ooalign{%
    $\m@th#1\cup$\cr
    \sbox0{$#1\cup$}%
    \dimen@=\ht0 %
    \sbox0{$\m@th#1\cdot$}%
    \advance\dimen@ by -\ht0 %
    \dimen@=.5\dimen@
    \hidewidth\raise\dimen@\box0\hidewidth
  }%
}

\providecommand*{\bigcupdot}{%
  \mathop{%
    \vphantom{\bigcup}%
    \mathpalette\@bigcupdot{}%
  }%
}
\newcommand*{\@bigcupdot}[2]{%
  \ooalign{%
    $\m@th#1\bigcup$\cr
    \sbox0{$#1\bigcup$}%
    \dimen@=\ht0 %
    \advance\dimen@ by -\dp0 %
    \sbox0{\scalebox{2}{$\m@th#1\cdot$}}%
    \advance\dimen@ by -\ht0 %
    \dimen@=.5\dimen@
    \hidewidth\raise\dimen@\box0\hidewidth
  }%
}
\makeatother

%% file: abs.tex
\begin{abstract}
  Multi-label classification is becoming increasingly ubiquitous, 
  but not much attention has been paid to interpretability. 
  In this paper, 
  we develop a multi-label classifier that can be re\-pre\-sent\-ed as a concise set of simple ``if-then'' rules, and thus, it offers better interpretability compared to black-box models. 
  Notably, our method is able to find a small set of relevant patterns 
  that lead to accurate multi-label classification, 
  while  existing rule-based classifiers are myopic and wasteful in searching rules, 
  requiring a large number of rules to achieve high accuracy.   
  In particular, 
  we formulate the problem of choosing multi-label rules to maximize a target function,  
  which considers not only discrimination ability with respect to labels, but also diversity. 
  Accounting for diversity helps to avoid redundancy, and thus, 
  to control the number of rules in the solution set.  
  To tackle the said maximization problem we propose a 2-approximation algorithm, 
  which relies on a novel technique to sample high-quality rules.  
  In addition to our theoretical analysis, 
  we provide a thorough experimental evaluation, 
  which indicates that our approach offers a trade-off between predictive performance and interpretability that is unmatched in previous work. 
\end{abstract}

%% file: intro.tex
\section{Introduction}
\label{sec:intro}

Machine-learning algorithms are nowadays being used in almost every domain. 
While such algorithms are known to perform well in many tasks, they are often used as ``black-boxes,'' i.e., the decision processes involved are too complex for humans to interpret.
The lack of interpretability limits considerably the level of trust humans put in machine-learning algorithms and thus,
poses a barrier for the wide adoption of machine-learning techniques in the real world. 
In an attempt to overcome this barrier, 
interpretable and explainable machine learning have recently emerged 
as increasingly prominent topics.
In the standard classification setting, the goal is to learn a classifier 
that accurately maps data points to two or more \emph{mutually exclusive} classes.

In this paper, we focus on a different setting,
namely, \emph{multi-label classification}. 
In contrast to the standard setting, 
in multi-label classification, 
a point can be associated with more than one class at the same time.
Though multi-label classification has been extensively studied, 
the main focus is still on improving predictive performance~\cite{tidake2018multi}.  %
Significantly less attention has been paid to interpretability aspects.

Classification rules, due to their simple structure, 
are gaining popularity in interpretable multi-label classification literature.
In rule-based approaches, the goal is to learn a set of rules 
that captures the most prominent patterns between features and labels in the data.
A rule usually takes the form ``\{set of predicates\} $\rightarrow$ \{set of labels\}.'' 
For a given data point, 
a rule would predict the associated labels to be present, 
if all the predicates in the rule evaluate to true.
Due to the structural simplicity of rules, 
classifiers based on a set of rules are generally considered more interpretable 
than other types of classifiers, such as neural networks or even decision trees.

The research question that we bring forward is whether 
we can design rule-based multi-label classification methods 
that are both accurate and interpretable.
\Boomer~\cite{rapp2020learning, rapp2021gradient}, 
a recently-proposed rule-based classifier based on gradient boosting, 
gives promising results in accuracy.
However, despite being a rule-based approach, its interpretability  
is limited
due to producing a set of rules that is both \emph{too large} and \emph{redundant}.

\begin{figure}[t]
	\centering
	\includegraphics[width=0.7\columnwidth, keepaspectratio]{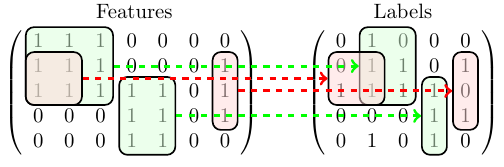}
	\vspace*{-0.15cm}\caption{\label{fig:example_rule}
          Illustration of the concept of our approach for multi-label rule selection.
          A toy dataset is visualized as a feature matrix and a label matrix.
          Four rules are shown as colored regions.
          Regions covered by the same rule are connected by a dashed arrow.
          The rules in green are chosen because of accuracy, generality, and diversity.
          The rules in red are discarded.}
\end{figure}

In this work, we propose \ouralg, a rule-based  method that significantly improves
over the state-of-the-art \Boomer.
The improvement is due to
(1) reducing rule redundancy, 
which is achieved by incorporating a term in our objective 
that penalizes for rule overlap,
and (2) explicitly limiting the complexity of rules 
via a suite of novel sampling schemes. 
As a result, our method produces a concise set of interpretable rules. 
An illustration of the concept of our approach 
is given in Fig.~\ref{fig:example_rule}.

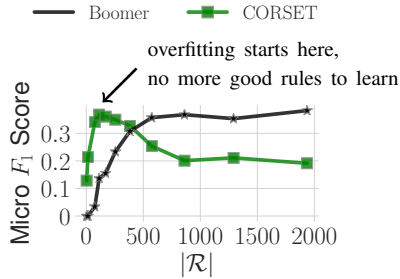
\begin{figure}[t]
  \centering
  \begin{tikzpicture}
    \node[draw=white, anchor=center] at (0, 2.2) {\includegraphics[width=0.6\columnwidth, height = 0.4\textheight, keepaspectratio]{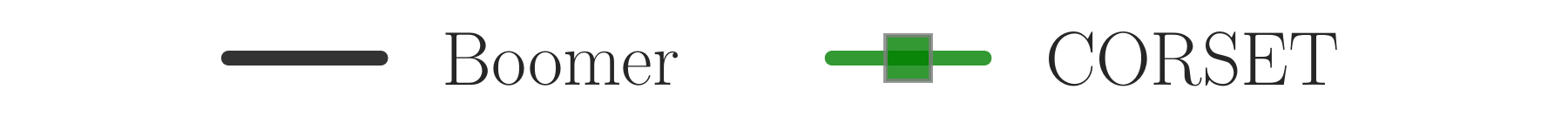}};
    \node[draw=white, anchor=center] at (0, 0) {\includegraphics[width=0.5\columnwidth, height = 0.5\textheight, keepaspectratio]{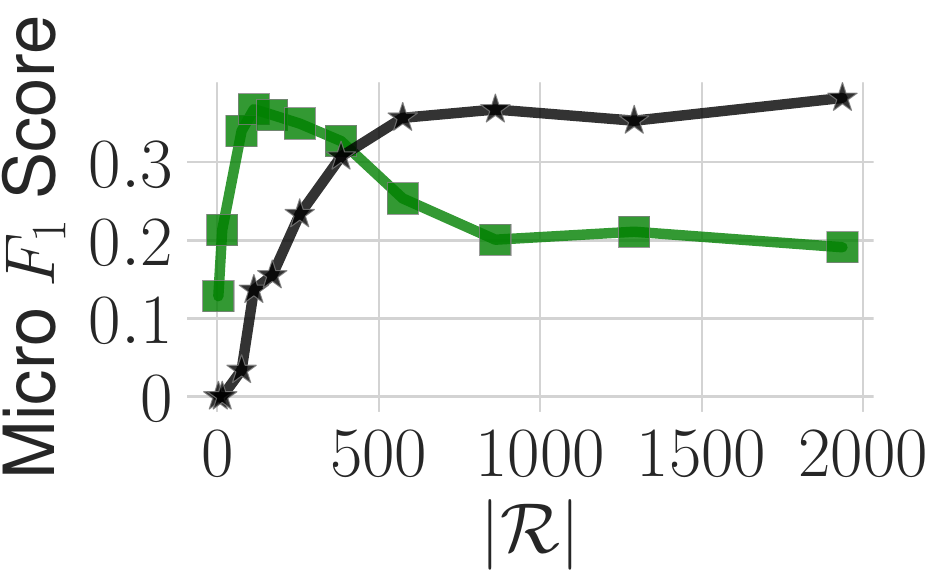}};

    \node[draw=white, anchor=west, text width=4cm]  at (-0.5, 1.7) {\footnotesize overfitting starts here,};
    \node[draw=white, anchor=west, text width=4cm]   at (-0.5, 1.32) {\footnotesize no more good rules to learn};
    \node[inner sep=0pt] (comment) at (-0.5, 1.5) {};
    \node[inner sep=0pt] (peak-point) at (-1, 1.) {};
    \draw[->, line width=1pt] (comment) -- (peak-point);
  \end{tikzpicture}
	\vspace*{-0.25cm}
	\caption{\label{fig:first_plot} 
	Micro $F_1$, as a function of number of rules
	for \ouralg vs.\ \Boomer
	in the \bibtex dataset. 
	}
\end{figure}

\begin{figure}[t]
\begin{tabular}{c}
	\includegraphics[width=1.\columnwidth, height = 0.2\textheight, keepaspectratio]%
	{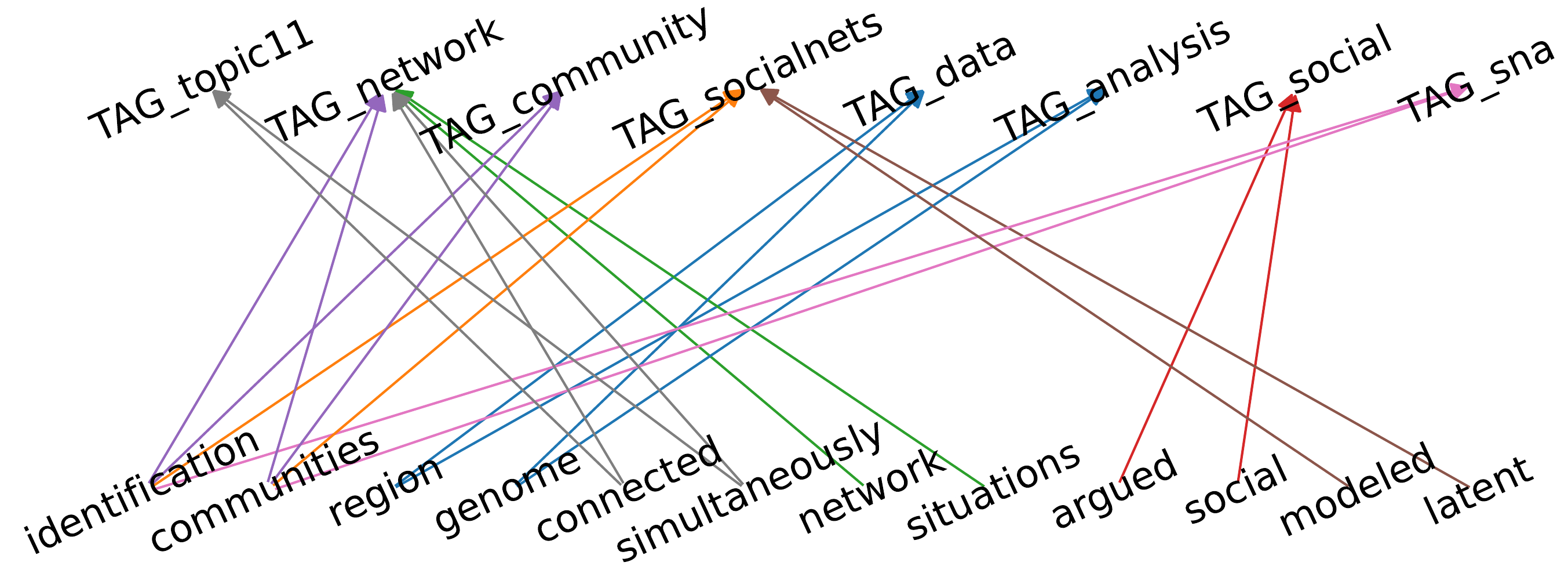}\\
	\includegraphics[width=1.\columnwidth, height = 1.\textheight, keepaspectratio]%
	{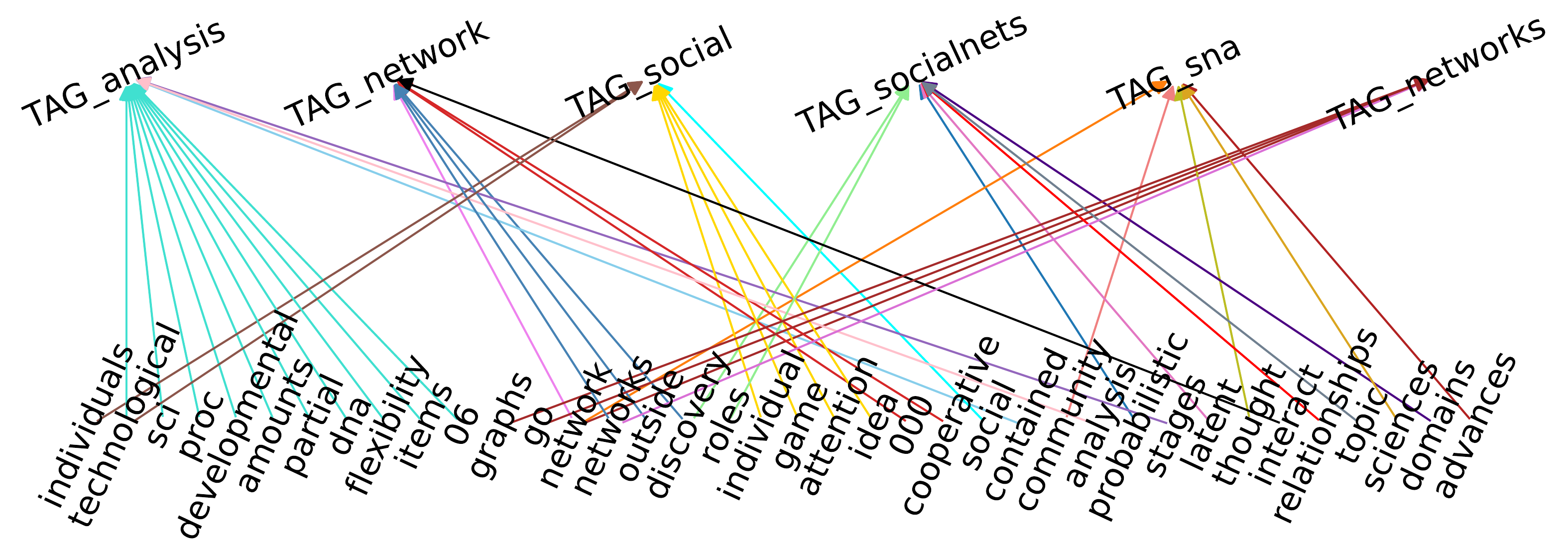}
	\end{tabular}
	\vspace*{-0.15cm}\caption{\label{fig:running_example} 
	An example set of rules returned by our algorithm (top) and \Boomer (bottom)
	on the \bibtex dataset. 
	We depict all rules for the set of labels 
	\{\textsf{\small SNA}, \textsf{\small socialnets}, 
	\textsf{\small social}, \textsf{\small networks}, 
	\textsf{\small analysis}\}.
	}
\end{figure}

\spara{Example.}
To illustrate the improvement of \ouralg over \Boomer, 
we consider as an example the \bibtex dataset, 
where each data point represents a scientific article, 
with bag-of-words as features and topics as labels.
We first consider predictive performance 
as a function of the number of rules.
In Fig.~\ref{fig:first_plot}, 
we show the (micro-averaged) balanced $F_1$ scores,
a popular measure for multi-label classification used throughout this paper,
for both \ouralg and \Boomer.
Due to the conciseness of its learned rules, 
\ouralg achieves a score close to $0.36$ with about 100 rules, 
whereas \Boomer needs over $800$ rules to achieve similar performance.
Note that \ouralg's performance starts to drop after about 100 rules, 
as there are no more good rules to learn.
The drop indicates overfitting, which can be addressed by standard methods, e.g., cross validation. 
In addition, Fig.~\ref{fig:running_example} 
demonstrates the conciseness of the rules found by \ouralg 
vs.\ the ones by \Boomer.
Here, we show a subset of rules as a bipartite graph, 
where nodes at the top represent labels 
and nodes at the bottom represent the predicates (features). 
Rules are represented by colors and two nodes are connected if they are part of 
the same rule. 
\ouralg uses fewer rules than \Boomer and rules tend to contain fewer predicates, 
resulting in a sparser graph.

Concretely, in this work we make the following contributions.  
\begin{itemize}
	\item We frame the problem of learning concise rule sets as an optimization problem. %
          The problem is \np-hard and our proposed algorithm \ouralg, given a set of rule candidates, achieves an approximation ratio of $2$. %
	
	\item The performance of \ouralg depends on the quality of the candidate rules.
          To find good rules efficiently, 
          we design a suite of fast sampling algorithms
          with probabilistic guarantees as well as an effective heuristic.
	
	\item %
           Our experiments show that \ouralg achieves competitive predictive performance compared to the state-of-the-art, 
           while offering significantly better interpretability. 
\end{itemize}

The rest of this paper is organized as follows. Section \ref{section:related-work} discusses related work. 
Section \ref{sec:prob} formalizes the problem we consider. Section \ref{sec:CMRS} illustrates \ouralgorithm, omitting the details of the rule-sampling algorithms it relies on, which are described in Section \ref{sec:sampling} and \ref{sec:enhancement}. 
Afterwards, Section \ref{sec:complexity} analyses the complexity of \ouralgorithm and finally Section \ref{sec:exp} presents a thorough experimental evaluation of \ouralgorithm. 

%% file: related.tex
\section{Related Work}
\label{section:related-work}

\para{Multi-label classification.}
In multi-label classification
the goal is to learn a function that maps input points 
to one or more predefined categories. 
For instance, a song can be associated with multiple music genres. 
A plethora of algorithms have been proposed for this problem;
interested readers may refer to a recent survey~\cite{tidake2018multi}. %
The simplest approaches for multi-label classification 
are the so-called transformation methods, 
which convert the original problem into multiple single-label classification problems.
The main drawback of these approaches is that they fail to capture label correlations. 
To overcome this issue, label power-set approaches map each distinct set of labels to a unique meta-label, 
which serves as target label for a single-label classifier.  
Clearly, these approaches do not scale with the number of labels and 
the pruned problem transformation method~\cite{read2008pruned} 
has been proposed as a remedy.
Another line of research focuses on designing ad-hoc multi-label classification methods 
by extending existing single-label algorithms.
Examples include adaption from support vector machines~\cite{elisseeff2001kernel}, 
$k$-nearest neighbor classifiers~\cite{zhang2007ml}, and perceptrons~\cite{crammer2003family}.

\spara{Interpretable machine learning.}
There is no agreed formal definition of interpretability, but 
it can be loosely defined as the degree to which a human can understand the cause of a 
decision~\cite{miller2019explanation}. 
Broadly speaking, interpretability in machine learning can be achieved 
by constraining the complexity of algorithms 
so that the process behind the decision of the algorithm is understandable to humans.
A related topic is explainable machine learning, 
where the goal is to provide explanations to the predictions of black-box models.

\spara{Rule-based approaches to single-label classification.}
Research in interpretable machine learning has boomed in the last years.
Rule-based (or associative) approaches 
have shown promising potential, because decisions are driven 
by a simple set of ``if-then'' rules.
Liu et al.~\cite{liu1998integrating} are among the first to investigate 
association rule mining for single-label classification tasks, followed by extensions such as
MCAR~\cite{thabtah2005mcar} and ADA~\cite{wang2011approach}.
These approaches are conceptually similar, but differ in their methodologies for rule learning, ranking, pruning, and prediction. 

\spara{Concise rule sets.} 
Our work pursues for the first time the goal of designing a multi-label associative classifier 
for achieving a given classification performance with the smallest possible number of rules. 
A similar objective has been recently considered in the context of single-label classification.
In particular, Zhang et al.~\cite{zhang2020diverse} 
frame the problem of learning a set of classification rules as an optimization problem 
with an objective function combining rule quality and diversity. 
A $2$-approximation algorithm is then proposed to solve this problem,  
which relies on existing frameworks for max-sum diversification and pattern sampling.  
In this paper, we investigate how to extend these ideas to the 
multi-label classification setting. 
The problem of controlling the number of rules has also been studied for rule boosting, 
where learned rules are combined additively~\cite{boley2021better}. 
An extension to multi-label classification represents a possible direction of future work. 
In addition to the number of rules, conciseness of a rule set, and thus interpretability, has been defined in terms of number of conditions~\cite{ghosh2022efficient} as well as of the Minimum Description Length principle~\cite{fischer2019sets}.

\spara{Rule-based approaches to multi-label classification.} In general, adaptation from the single-label to the multi-label setting is not trivial 
and while single-label associative classification has been studied extensively,  
relatively few attempts have been
made for associative multi-label classification. 
In an early work, Thabtah et al.~\cite{thabtah2004mmac} propose a label ranking-based assignment method. 
More recently new approaches have been developed, and 
\seco~\cite{yk:Relaxed-Pruning} and \Boomer~\cite{rapp2020learning} 
are state-of-the-art in the current literature of rule-based multi-label classification.
The main limitation of the existing works, 
addressed in our paper, 
is that they use a very large set of highly redundant rules, 
which hinders interpretability. 
We compare our method against \seco~\cite{yk:Relaxed-Pruning} and 
\Boomer~\cite{rapp2020learning} in Section~\ref{sec:exp}.

\spara{Pattern sampling.}
Association pattern discovery is challenging due to 
the prohibitive size of the pattern space. 
This challenge is inherited by rule-based classifiers. 
To avoid exhaustively searching the pattern space, 
efficient pattern-sampling methods 
have been proposed~\cite{boley2011direct,boley2012linear}. 
In this work we extend these sampling methods 
to efficiently find high-quality candidate multi-label rules, 
as discussed in detail in Section \ref{sec:sampling}.

%% file: problem_formulation.tex
\section{Problem Statement}
\label{sec:prob}

At a high level, our objective is to capture the relevant patterns in the data 
that best discriminate a set of labels and are as concise as possible.
Next we formally define the problem. 

\subsection{Preliminaries}
\label{sec:prelim}

We denote sets and multi\-sets by upper\-case letters e.g., $X$.
For a finite ${X}$, we denote by $\powerset\pr{X}$ its power set. 
We consider a binary \emph{dataset} \ds over a \emph{feature set} \featspace 
and a \emph{label set} \labelspace.
The dataset $\ds$ is a set of \emph{data records}, $\dr_1, \ldots, \dr_{\numpts}$.
A data record $\dr=(\feats, \labels)$  consists of a set of features $\feats \subseteq \featspace$ and a set of labels $\labels \subseteq \labelspace$. 
We denote by $\featsD$ and $\labelsD$ the feature set and label set of \dr, respectively. 
Furthermore, we denote by $|\featspace|$ and $|\labelspace|$ the dimensions of the feature and label space, respectively, and we denote by $|\dataset|$ the total number of data records. 
We use $\|\featspace\|$ and $\|\labelspace\|$ to refer to the total number of 
feature and label occurrences over all data records. 

In multi-label classification, 
the goal is to learn a function mapping as accurately as possible 
the features $\featsD$ to one or more labels \labelsD. 
We use mappings consisting of \emph{conjunctive rules}.
A {conjunctive rule} \ruledef consists of a non-empty feature set~$\rhead$ 
(called \emph{head}) and a non-empty label set~$\rtail$ (called \emph{tail}).
The head $\rhead$ can be viewed as a predicate 
$\rhead: \set{0, 1}^{\abs{\featspace}} \rightarrow \set {\text{true}, \text{false}}$,
which states whether an instance $\feats$ contains all the features in $\rhead$.
If the predicate evaluates to $\text{true}$ for some instance,
the tail $\rtail$ of $\crule$ specifies that labels $\rtail$ should be predicted as present. 

We say that a head $\rhead$ \emph{matches} a data record $\dr$ if $\rhead \subseteq \featsD$. 
Similarly, a tail $\rtail$ matches $\dr$ if $\rtail \subseteq \labelsD$.
We say that a rule $\crule$ \textit{covers} a data record $\dr$ if $\headr \subseteq \featsD$ and
similarly $\crule$ matches a data record $\dr$ if both $\headr$ and $\tailr$  match $\dr$.
For a dataset~\ds, we denote the \emph{support set} of 
$X \in \{\rhead, \rtail, \crule \}$ by:
\[
   \dsX=\set{\dr \in \dataset \mid X \text{ matches } \dr}.
\]
The \emph{space of all possible rules} we consider is 
$\rulespace = \powerset\pr{\featspace} \times \powerset\pr{\labelspace}$, 
i.e., the Cartesian product of the power set of the feature set and 
the power set of the label set.

\subsection{Problem formulation}
\label{sec:prob_form}

We want to discover rules that are accurate and general, 
but also sufficiently different from each other.
To capture this trade-off,  
we design an objective function that consists of 
a \emph{quality term} $\quality: \rulespace \rightarrow \real$ 
measuring the accuracy and generality of a single rule, and
a \emph{diversity term} $\dist: \rulespace \times \rulespace \rightarrow \real$ 
measuring the distance between pairs of rules.

\spara{Quality term.}
Given a rule \crule and a set of rules \ruleset, the quality $\quality(\crule; \ruleset)$ 
of $\crule$ with respect to \ruleset
is the product of two values:
the \emph{uncovered area} $\relarea{\crule}$, capturing the generality of \crule 
with respect to \ruleset,
and its \emph{adjusted accuracy} $\disc(\crule)$,
\[
  \quality(\crule; \ruleset) = \relarea{\crule} \cdot \disc(\crule).
\]
Next we describe these two functions.
To capture generality, we first define the \emph{coverage} of $\crule$ as:
\begin{equation}
\label{eq:coverage}
\cov\pr{\crule} = \set{\pr{i, k} \mid \crule \matches \dr_i \in \ds \text{ and } k \in \rtail}.
\end{equation}
In other words, the coverage of a rule is the set of label occurrences it matches in a dataset.
To incorporate what is already covered by a set of selected rules \ruleset, 
we define \textit{uncovered area} of \crule with respect to \ruleset as

\begin{equation}
\label{eq:relarea}
\relarea{\crule} = \abs{\cov\pr{\crule} \setminus \bigcup\limits_{\crule \in \ruleset} \cov\pr{\crule}}, 
\end{equation}
that is, the size of covered label occurrences by \crule after excluding those already covered by \ruleset.
Thus, a rule \crule is considered general 
with respect to \ruleset if $\relarea{\crule}$ is large. 

Before introducing the {adjusted-accuracy} function, 
we need some additional notation. 
Data records whose labels contain \rtail are said to be positive with respect to \rtail,
whilst the remaining ones are negative. More formally,
a tail \rtail bi-partitions a dataset \dataset into two disjoint sets:
a set of \textit{positive data records} 
$\positives = \{ \dr  \in \dataset \mid \rtail \subseteq \labelsD\}$
and a set of \textit{negative data records} 
$\negatives = \{ \dr \in \dataset \mid \rtail \nsubseteq \labelsD\}$.
Given a rule \crule,
let $\PdsR = \frac{\abs{\dsR}}{\abs{\dsH}}$ be the \textit{precision} of $\crule$
and $\Pds = \frac{\abs{\dsT}}{\abs{\ds}}$ is the \emph{base rate} of \rtail in \ds.
We denote the corresponding binomial distributions as 
$\bindist\pr{\PdsR}$ and $\bindist\pr{\Pds}$, respectively. 
Then the \textit{adjusted accuracy} of \crule is defined as:
\begin{equation}
  \label{eq:disc}  
\disc\pr{\crule} = \indicator\pr{\crule} \cdot \Dkl\pr{\bindist\pr{\PdsR}\mid\mid\bindist\pr{\Pds}}, 
\end{equation}
where $\indicator\pr{\crule}$ is 1 if $\PdsR > \Pds$ and 0 otherwise, 
and $\Dkl\pr{\cdot \mid\mid \cdot}$
is the KL divergence between two probability distributions. 
The underlying intuition is that 
if the precision of a rule is below its base rate, 
it is useless, and receives a zero score.
If instead the precision of a rule is larger than the base rate, 
the higher the precision is, the larger the score. 

\spara{Diversity term.}
We measure the distance between two rules by how much their coverages overlap.
Formally, given two  rules $\crule_1$ and $\crule_2$, their distance  is defined as
\[
  \dist\pr{\crule_1, \crule_2} = 1 - \frac{\abs{\cov\pr{\crule_1} \cap \cov\pr{\crule_2}}}{\abs{\cov\pr{\crule_1} \cup \cov\pr{\crule_2}}},
\]
which is the \emph{Jaccard distance} 
between $\cov\pr{\crule_1}$ and $\cov\pr{\crule_2}$.

\spara{Problem definition.}
We frame the learning problem as a combinatorial optimization problem with budget constraint,
where we set a budget on the maximum number of rules to discover,
and rules should be selected to maximize a linear combination 
of the quality and diversity term.

\begin{problem}
   \label{pr:main}
   Given a dataset \dsdef, a budget $\budget\in\integer$, 
   a space of rules $\samplespace \subseteq \rulespace$,  
   and a parameter $\lambda \in \real_+$, 
   find a set of $\budget$ rules 
   $\ruleset = \set{\crule_1, \ldots, \crule_\budget} \subseteq \samplespace$, 
   to maximize the following objective
   \begin{equation}
     \label{eq:obj}
     \obj\pr{\ruleset} = \sum\limits_{\crule \in \ruleset} \quality\pr{\crule; \ruleset \setminus \set{\crule}} + \lambda \sum\limits_{\crule_i, \crule_j \in \ruleset, i \neq j} \dist(\crule_i, \crule_j).
   \end{equation}
\end{problem}
This problem is known to be NP-hard~\cite{borodin2012max}.
In the next section, we present a greedy algorithm which finds a solution to Problem~\ref{pr:main} with an approximation factor of $2$, provided that the space of rules  $\samplespace$ can be visited in polynomial time.

%% file: algorithm.tex
\section{{\small CORSET} Learning Algorithm}
\label{sec:CMRS}

In this section,
we present a meta algorithm named \ouralgorithm 
(\emph{\underline{co}ncise \underline{r}ule \underline{set}}) 
for Problem~\ref{pr:main}.
\ouralg greedily picks one rule at a time 
from a pool of candidate rules, 
so as to maximize the marginal gain for the objective in~\eqref{eq:obj},
i.e.,
\[
\obj'\pr{\ruleset \cup \set{\crule}} - \obj'\pr{\ruleset} = \frac{1}{2} \quality\pr{\crule; \ruleset} + \lambda \sum\limits_{\crule_j \in \ruleset} \dist(\crule, \crule_j).
\]
The candidate rules are generated by a procedure called \samplecandalg.
The effectiveness of \samplecandalg 
heavily affects the predictive performance of the classifier.
The goal is to sample high-quality rules in terms of generality, diversity and accuracy.
This is a challenging goal since the size of rule space is exponential~\cite{furnkranz2012foundations} 
and \samplecandalg should therefore avoid exploring the whole space. 
We defer the description of the candidate generation to 
Sections~\ref{sec:sampling} and~\ref{sec:enhancement}.

For now, we focus on the description of the main algorithm.
\ouralg maintains a set of selected rules, $\ruleset$, which is initially empty. 
At each iteration, \ouralg considers a pool of candidate rules 
generated by \samplecandalg.
Within this pool, the rule  $\crule^*$  maximizing the marginal gain of~\eqref{eq:obj} 
with respect to \ruleset is selected and added to $\ruleset$. 
The process stops when the proportion of labels in \labelspace predicted by $\ruleset \cup {\crule^*}$ and not by $\ruleset$ falls below a user-specified tolerance level $\tol$.

To ensure the aforementioned approximation guarantee,
we need to obtain a new set of rules $\ruleset'$ by repeating the greedy procedure a second time, 
over the full set of candidates rules, 
because using a different candidate set at each iteration does not offer a guarantee. 
As in practice $\ruleset'$ is not necessarily better than $\ruleset$, 
we return as solution the set that yields the largest objective function value, between the two.

The pseudocode of  \ouralg is shown in Algorithm~\ref{alg:CMRS}.
Note that \samplecandalg receives as input the current rule set~$\ruleset$, 
so as to generate rules different from $\ruleset$.
The solution is guaranteed to be within a constant factor of the optimal solution.

\smallskip
\begin{proposition}
\label{proposition:approx}
For a fixed pool of candidate rules, 
\ouralg is a $2$-approximation algorithm for Problem~\ref{pr:main}. 
\end{proposition}
\smallskip

As shown by Borodin et al.~\cite{borodin2012max}, 
the approximation factor is guaranteed by the properties of the objective function, 
namely by the \emph{submodularity} of the quality function 
and the fact that the Jaccard distance is a \emph{metric}. 
Proof is given in Appendix. 

\begin{algorithm}[t]
	\SetAlgoLined
	\KwData{ data \dataset,  tolerance $\tol$}
	\KwResult{a set of multi-label classification rules \ruleset.}
	\caption{
          The \ouralgorithm  algorithm.
		\label{alg:CMRS}
	}  
	
	\ruleset, $\ruleset'$ $\leftarrow$ $\emptyset$, $\candidaterules \leftarrow \emptyset$\;
	
	\While{$ c > \tol $}{
		
		\currentrules $\leftarrow$ $\samplecandalg(\ruleset)$\; 
		
		 $\candidaterules \leftarrow \candidaterules \cup \currentrules$\; 
		
                 $\crule^* \leftarrow \argmax_{\crule \in \currentrules} \spr{\obj'\pr{\ruleset \cup \set{\crule}} - \obj'\pr{\ruleset}}$\; 
		
                 $c \leftarrow  \frac{ \sum_{\dr \text{ matched by } \crule^*} | \labelsD \cap T_{\crule^*} \setminus \cup_{\crule \in \ruleset \mid \crule \text{ matches } \dr} \tailr  |}{\|\labelspace\|}$\;

		$\ruleset \leftarrow \ruleset \cup \crule^*$\;

	}
	
	\For{$i =1,..|\ruleset|$}{
		
          $\crule^* \leftarrow \argmax_{\crule \in \candidaterules} \spr{\obj'\pr{\ruleset^{'} \cup \set{\crule}} - \obj'\pr{\ruleset^{'}}}$\; 
		
		$\ruleset' \leftarrow \ruleset' \cup \crule^*$\; 
		
	}
	
	\lIf{$\obj(\ruleset')  > \obj(\ruleset)$}{\Return{$\ruleset'$}}
	\lElse{\Return{$\ruleset$}}
	\end{algorithm}

\spara{Prediction.}
At prediction time,
given the set of selected rules~\ruleset,
we return the set of predicted labels for an instance $\feats$ as
$\bigcup_{\crule \in \ruleset \mid \headr \subseteq \feats }  \tailr$,
that is, the union of tails of rules such that $\feats$ evaluates to $\text{true}$ for the head predicate.

%% file: samp.tex
\section{Rule Sampling}
\label{sec:sampling}

In the next two sections, 
we present the main contribution of our work, 
a suite of rule-sampling algorithms used by $\samplecandalg$.
In this section, we first describe the technical basis of our proposal, 
then we formulate our sampling problem, and present our algorithms for it.
In the next section, 
we discuss some important limitations of the proposed sampling method 
and describe practical enhancements.
    
\subsection{Background: Two-stage pattern sampling}
\label{sec:sampling_background}

Our sampling scheme builds on the pattern-sampling algorithms proposed by 
Boley et al.~\cite{boley2011direct, boley2012linear}. 
These algorithms allow us to sample patterns according to a target distribution 
over the pattern space, without the need of exhaustive enumeration. 
The target distribution reflects a measure of interestingness for the patterns.
Example measures include support, area, and, if the data are labelled, discriminativity. 
Sampling algorithms for a variety of measures share a two-stage structure,
whilst the details
depend on the measure under consideration.

The key insight brought by Boley et al.~\cite{boley2011direct, boley2012linear} is that \textit{random experiments reveal frequent events}.
We use sampling by support and area for illustration.
Consider a dataset $\ds=\set{\dr_1, \ldots, \dr_{\numpts}}$ over a finite ground set $\groundset$, 
with $\dr \subseteq \groundset$ for each $\dr \in \ds$.
Consider the problem of sampling an itemset (pattern) $F \subseteq \groundset$ with probability proportional to its \textit{support} $\qsupp\pr{F}=\abs{\dr\spr{F}}$.

For each $\dr \in \ds$, the set of itemsets including $\dr$ in their support is $\powerset\pr{\dr}$.
It can be shown that
sampling  an itemset $F$ uniformly from $\bigcupdot_{\dr \in \ds} \powerset\pr{\dr}$, 
where $\bigcupdot$ denotes the union operator of multi-sets,
is the same as sampling $F$ according to $\abs{\dr{\spr{F}}}$.
To avoid materializing $\bigcupdot_{\dr \in \ds} \powerset\pr{\dr}$, 
Boley et al.\ use a two-step procedure:
\begin{enumerate}
\item[1.] sample a data record \dr with probability proportional to the weight 
					$\wgt\!\pr{\dr}= \sum_{F \in \powerset\pr{\dr}} 1 = 2^{\abs{\dr}}$. 
\item[2.] sample an itemset $F$ uniformly from $\powerset\pr{\dr}$.
\end{enumerate}
To sample from the ``area'' distribution
$\qarea\pr{F} = \abs{F}\abs{\dr\spr{F}}$, 
the above procedure is changed as follows:
$\wgt\!\pr{\dr} = \sum_{F \in \powerset\pr{\dr}} F = \abs{\dr} 2^{\abs{\dr} - 1}$,
and then sample $F$ with weight $\abs{F}$ from $\powerset\pr{\dr}$.

The two-stage sampling idea can be generalized to a number of other measures. Some of them, 
such as discriminativity, which we use later, 
require sampling tuples of data records rather than a single one in the first stage.

Next we describe two sampling distributions and the 
corresponding sampling algorithms for our objective.
The first distribution is a generalization of the area function 
(not discussed by Boley et al.~\cite{boley2011direct, boley2012linear}) 
and is used for tail sampling.
The second distribution is discriminativity and is used for head sampling.
For the latter, 
we propose an improved sampling algorithm, 
which is faster than the original version~\cite{boley2012linear}.

\subsection{Sampling objectives}
\label{sec:sampling_objective}

Our rule sampling objective
is a product of two values, 
reflecting the generality of a rule \ruledef
given the current set of rules \ruleset, 
and its discriminative power:
\begin{equation}
\label{eq:sampling-obj}
\proba{\rhead, \rtail}  \propto  \wgt(\rhead, \rtail; \ruleset) = \qdist\pr{\rhead; \rtail} \cdot \relarea{\rtail}.
\end{equation}
Note that the uncovered area function in~\ref{eq:relarea} generalizes to tails, i.e.,
$\relarea{\rtail} = \abs{\cov\pr{\rtail} \setminus \bigcup_{\rtail^{'} \in \ruleset} \cov\,(\rtail^{'})}$ and 
$\cov\pr{\rtail} = \set{\pr{i, k} \mid \rtail \matches \dr_i \in \ds \text{ and } k \in \rtail}$.

For $\qdist$ we choose the discriminativity measure studied 
by Boley et al.~\cite{boley2011direct}, 
which permits sampling in polynomial time.

Given a tail $\rtail \subseteq \labelspace$,
the \textit{discriminativity} of $\rhead$ is defined as
\begin{equation}
\label{eq:sampling-disc}
q_\text{disc}\!\pr{\rhead; \rtail} = \abs{\positives\spr{\rhead}}\; \abs{\negatives \setminus \negatives \spr{\rhead}}.
\end{equation}
The goal is to sample heads that have as large support as possible in \positives
and as small support as possible in \negatives.

To sample from distribution \eqref{eq:sampling-obj}, 
we use the following steps:
\begin{enumerate}
	\item[1.] sample \rtail with probability proportional to $\relarea{\rtail}$;
	\item[2.] sample \rhead with probability proportional to $\qdist\pr{\rhead; \rtail}$.
\end{enumerate}
We explain each sampling step next.

\subsection{Tail sampling}
\label{sec:tail-sampling}

To sample from $\relarea{\rtail}$, 
we apply a similar two-step sampling procedure as in Boley et al.~\cite{boley2011direct}:
we first sample a data record \dr with probability proportional 
to its weight $\wgtR{\dr}$ 
and then sample \rtail from \dr.
The function $\relarea{\rtail}$ is a generalization of the area function considered in Boley et al.~\cite{boley2011direct}.
Adapting the original algorithm to our case requires to design a weight function $\wgtR{\cdot}$ appropriate for our target.
To define  $\wgtR{\cdot}$, a few new definitions are needed. 
Given a rule \crule and a data record \dr,
the \emph{\dr-specific coverage} of \crule is defined to be
\begin{equation}
\label{eq:covD}
\covD\!\pr{\crule} = \begin{cases} 
	\labelsD \cap \tailr, &\text{ if } \crule \text{ matches } \dr, \\ 
	\emptyset, & \text{otherwise.}
\end{cases}
\end{equation}
Extending \dr-specific coverage to a rule set \ruleset, we have:
\begin{equation}
\label{eq:covDset}
\covD\!\pr{\ruleset} = \bigcup\limits_{\crule \in \ruleset} \covD\!\pr{\crule}. 
\end{equation}
Given a label set \rtail, its \textit{marginal coverage} with respect to \ruleset~is
\begin{equation}
\label{eq:covDtail}
\relcov{\rtail} = \pr{\labelsD \cap \rtail} \setminus \covD\!\pr{\ruleset},
\end{equation}
that is, the covered label occurrences in \dr by \rtail,
excluding those by \ruleset.
As a shortcut, we define  
$\relcovR = \relcov{\labelsD}$, i.e., the set of label occurrences in \dr not covered  by \ruleset.

The weight of a label set $\rtail$ on  a data record \dr is:
\begin{equation}
\label{eq:wgtR-tail}
\wgtR{\rtail, \dr} = \abs{\relcov{\rtail}}.
\end{equation}

We give a small example to illustrate these definitions:
\begin{center}
    \begin{tikzpicture}[scale=0.9,
      box/.style={draw=white},
      ]
      \node[box, anchor=center] at (0, 1.2) {$D: \featsD=\set{0, 1, 2, 3}, \;\labelsD=\set{a, b, c}$};
      \node[box, anchor=west] at (-3, 0.5) {$\crule_1: \set{0, 1} \rightarrow \set{a}$};
      \node[box, anchor=west] at (-3, -0.25) {$\crule_2: \set{1, 2} \rightarrow \set{a, b}$};
      \node[box, anchor=west] at (-3, -1) {$\crule_3: \set{2, 3} \rightarrow \set{a, c}$};

      \node[box, anchor=west] at (0.5, 0.2) {$\ruleset=\set{\crule_1, \crule_2, \crule_3}$};
      \node[box, anchor=west] at (0.75, -0.6) {$\rtail=\set{b, c}$};
    \end{tikzpicture}
\end{center}
For  $\crule_1, \crule_2, \crule_3$, 
the sets $\covD\!\pr{\cdot}$ are $\set{a}, \set{a, b}, \emptyset$, respectively.
Therefore, $\relcov{\rtail} = \set{c}$ and $\wgtR{\rtail, \dr} = 1$.

The intuition of the definition of $\wgtR{\rtail, \dr}$ 
is that $\rtail$ has large weight on $\dr$ 
if it contains many  label occurrences not covered by $\ruleset$.
Therefore, the weight of any data record \dr 
is simply the summation of the weights over all possible tails:
\begin{equation}
\label{eq:wgtRD}
\wgtR{\dr} = \sum_{\rtail \subseteq \labelsD} \wgtR{\rtail, \dr} = \abs{\relcovR} \;2^{\abs{\labelsD} - 1},
\end{equation}
where the second equality can be shown by simple algebra. 
Using these weights, we adapt the sampling algorithm of 
Boley et al.~\cite{boley2011direct, boley2012linear} 
as per Algorithm~\ref{alg:tailSampling}. 
\begin{algorithm}[t]
  \SetAlgoLined
  \KwData{
    a dataset \ds, weights \wgtR{\dr} (as in~\ref{eq:wgtRD}).
  }
  \KwResult{a tail $\rtail \subseteq \labelspace \textrm{ with } \rtail \sim \relarea{\rtail}$.}%
  \caption{Two-stage tail sampling.
    \label{alg:tailSampling}
  }  
  draw  $\dr \sim \wgtR{\dr}$\; %
  \Return{ $\rtail \sim \wgtR{\rtail, \dr}$}
\end{algorithm}

By a similar proof technique as in Boley et al.~\cite{boley2011direct}, we have:

\smallskip
\begin{proposition}
  Algorithm \ref{alg:tailSampling} returns $\rtail \sim \relarea{\rtail}$. 
\end{proposition}
\smallskip
   
The proof is provided in Appendix.

\subsection{Head sampling} 
\label{sec:head-sampling}

After a tail \rtail is sampled,
we sample \rhead according to 
$\qdist\pr{\rhead; \rtail} = q_\text{disc}\!\pr{\rhead; \rtail}$, 
from~\eqref{eq:sampling-disc}.
The two-stage sampling scheme by Boley et al.
can be applied for this case.
In contrast to the previous cases,
the weight function
is defined on \textit{pairs} of data records:
\begin{equation}\label{eq:weights_discr}
\wgt(\positiveD, \negativeD) = 2^{\abs{\positiveD} }- 2^{\abs{\positiveD \cap \negativeD}} - \abs{\positiveD \setminus \negativeD},
\end{equation}
where $\positiveD \in \positives$ and $\negativeD \in \negatives$, and 
$\abs{\positiveD}$ (resp.\ $\abs{\negativeD}$) denotes the number of features present 
in $\positiveD$ (resp.\ $\negativeD$). 
Thus, pre-computing the weights leads to \textit{quadratic} space complexity in $\abs{\ds}$, which limits the practicality of the sampling procedure.

The above limitation is addressed by Boley et al.~\cite{boley2012linear}  
using the technique of \emph{coupling from the past} (\cftp), 
which leads to \textit{linear} space complexity.
Unlike many Markov chain Monte Carlo (\mcmc) methods, 
\cftp can guarantee that samples are generated according to the target distribution.
It operates by simulating the Markov chain \emph{backwards} by sampling from a proposal distribution, until all states coalesce to the same unique state.
The main challenge of using \cftp is the design of the proposal distribution and 
the efficient monitoring of coalescence condition.

The proposal distribution should be
($i$) efficient to sample from; and 
($ii$) an appropriate approximation to the target distribution 
to obtain fast convergence.
Boley et al.~\cite{boley2012linear} devise a ``general-purpose'' proposal distribution, 
which works for all target distributions they consider.
For the case of discriminativity,
the proposal distribution is defined as
\begin{equation}
 \label{eq:boley-proposal}
  \propodist(\posD, \negD) = \sqrt{\wgtpos\!\pr{\posD} \cdot \wgtneg\!\pr{\negD}},
\end{equation}
where $\wgtpos\!\pr{\posD}=2^{\abs{\posD}} - \abs{\posD} - 1$ and $\wgtneg\!\pr{\negD}=2^{\abs{\featspace}} - 2^{\abs{\negD}} - \abs{\negD} - 1$.
Sampling from $\propodist\!\pr{\cdot, \cdot}$ 
can be done efficiently by sampling separately from 
$\wgtpos\!\pr{\cdot}$ and $\wgtneg\!\pr{\cdot}$.
However, we argue that the choice of $\propodist\!\pr{\cdot, \cdot}$ 
is not a good approximation of the target, 
and therefore it suffers from slow convergence. 
The reason is that when a data record is high-dimensional but sparse, 
as is often the case in multi-label classification, 
$\wgtneg\!\pr{\negD}$ and hence $\propodist\!\pr{\posD, \negD}$ 
grow exponentially with the number of features, 
making the acceptance probability extremely low and, as a consequence, 
convergence is extremely slow.  

To overcome the convergence issue,
we use a different proposal distribution better suited for our setting.
Our proposal is the same as in~\eqref{eq:boley-proposal}, 
except that $\wgtneg$ is defined as a uniform function,  
$\wgtneg\!\pr{\negD} = 1$ for all $\negD \in \negatives$ and the square root is removed.
An appealing property is that the new choice is a \textit{tight} upper bound of~\ref{eq:weights_discr},
therefore providing a better approximation to the original version.
Further, we empirically verify that using our proposal 
gives much faster convergence than using the one by Boley et al~\cite{boley2012linear}.

Head sampling is summarized in Algorithm~\ref{alg:headSampling}.
We first use \cftp 
(lines~\ref{head-sampl:cftp-start}-\ref{head-sampl:cftp-end}) 
to sample a pair $\pr{\posD, \negD}$.
Then we sample a head in line~\ref{head-sampl:pattern}.
We denote $u(\cdot)$ as the uniform distribution over a set.
For brevity, we use a boldface letter to denote a pair of records, e.g., $\tuple$.
We denote an empty pair by \nullobj, and 
define $\propodist(\nullobj) / \wgt(\nullobj) = 1$.

\begin{algorithm}[t]
	\SetAlgoLined
	\KwData{a dataset \ds, a tail $\rtail$, weights 
					$\wgtpos(\cdot)$ and $\wgtneg(\cdot)$.}
	\KwResult{a head $\rhead \in \featspace \textrm{ with } \rhead \sim \qdist\!\pr{\rhead; \rtail}$.}%
	\caption{Two-stage head sampling.\label{alg:headSampling}}
	initialize $i \leftarrow 1$, $\tuple \leftarrow \nullobj$\label{head-sampl:cftp-start}\; 
	\While{$\tuple = \nullobj$}
	{
		$i \gets i + 1$\;
		\For{$t = 2^i, \ldots, 0$} 
		{
                  draw $u_t \sim u(\spr{0,1})$ \text{ and } $\proposalstate \sim \propodist\!\pr{\proposalstate}$\;
                  \lIf{
                    $\mathbf{u}_t \leq \frac{\propodist(\tuple)\wgt(\proposalstate) }{\wgt(\tuple)\propodist(\proposalstate)}$
                  }{
                    $\tuple \gets \proposalstate$
                  }
                }
                \label{head-sampl:cftp-end}
              }
	\hspace*{-0.5mm}draw $\rhead_1 \sim u(\powerset(D^+ \setminus D^-) \setminus \emptyset),\rhead_2 \sim u(\powerset(D^+ \cap D^-))$\label{head-sampl:pattern}\;
	\Return{$\rhead = \rhead_1 \cup \rhead_2$}    
\end{algorithm}

%% file: enhanced_sampling.tex
\section{Enhancements to the Sampling Scheme}
\label{sec:enhancement}

\subsection{Limitations of the two-stage pattern-sampling framework}
While theoretically sound, in our setting, 
the two-stage sampling framework 
\cite{boley2011direct, boley2012linear} suffers from two limitations, 
as can be verified empirically.
First, we observe that most of the sampled rules are very specific, with very low support.
Second, rule interpretability is not explicitly considered.

\spara{Heavy-hitter problem for tail sampling.} 
Consider the tail sampling part.
Notice that the weight of a data record $\dr$ in~\eqref{eq:wgtRD} 
is exponential in $\labelsD$.
If there is a data record $\dr \in \ds$ whose $\abs{\labelsD}$ is moderately larger than the rest,
its weight dominates, making it very likely to be sampled in the first sampling step.
We refer to this issue as \textit{the heavy-hitter problem}.
For instance in \bibtex,
the largest label set of a data record \drOne %
contains $28$ labels while the second
largest contains $16$.
The probability of \drOne being sampled is $99.97\%$. %
A tail sampled from \drOne has an expected length of  $14.5$.
Empirically, tails of about this length match only a few data records.
Thus, most of the sampled tails have low support, 
hampering the goal of sampling general rules.

\spara{Heavy-hitter problem for head sampling.}
A similar issue arises in head sampling.
The weight function in~\eqref{eq:weights_discr} grows exponentially with $|D^+|$,
so that \cftp most likely returns the positive data records with the highest number of present features.
Therefore, sampled heads  %
tend to be very long and have small support (often $1$).
Thus, they may have high discriminativity but %
cannot generalize to unseen data.  

\spara{Tail interpretability.} %
Interpretability of tails is a central focus in our work. 
Nonetheless, in the original pattern-sampling algorithms~\cite{boley2011direct, boley2012linear} 
all elements in $\powerset\!\pr{\labels}$ are considered possible tails,
regardless of whether they are interpretable or not.
Some tails are sampled simply because labels in them co-occur frequently, 
rather than because they are truly interpretable.

The root of the above limitations is the enormous sample space under consideration, 
which we address next.
\subsection{Tail sampling under interpretable label space}\label{sec:surs}

We propose to restrict the label sample space to a much smaller sample space $\reducedss \subseteq \powerset\!\pr{\labelspace}$ designed to contain only in\-ter\-pret\-able label sets 
so as to mitigate the heavy-hitter problem. 
We call $\reducedss$ the \textit{interpretable label space}.
Before describing the construction of $\reducedss$,
we notice that pattern sampling  under any subspace of $\powerset\!\pr{\labelspace}$ is a slight generalization of the original sampling setting. 
Most importantly, the original sampling algorithms can be adapted to different sample spaces,
such as~\reducedss, 
while preserving probabilistic guarantees. 

In Algorithm~\ref{alg:sampling-red},
we describe a procedure for sampling by uncovered area under $\reducedss$.
The algorithm can be easily adapted for other sampling objectives e.g., discriminativity.
Compared to sampling under $\powerset\!\pr{\labelspace}$,
we require the extra step of determining the set $\invmap\spr{\dr}$  of patterns in $\reducedss$ 
contained by $\dr \in \ds$ and computing the weight for $\dr$ accordingly.

\begin{algorithm}[t]
	\SetAlgoLined
	\KwData{
          a dataset \ds, sample space $\reducedss$, and a rule set \ruleset
	}
	\KwResult{a tail $\rtail \in \reducedss \textrm{ with } \rtail \sim \relarea{\rtail} $.}%
	\caption{
          Tail sampling  under $\reducedss$ according to uncovered area.
          \label{alg:sampling-red}
        }%
        let $\invmap\spr{\dr} \leftarrow \set{S \in \reducedss \mid S \subseteq \labelsD}, \textrm{ for each } \dr \in \ds$\label{sampling-red:preproc}\;
        let $\wgt\pr{\dr} \leftarrow \sum_{S \in \invmap\spr{\dr}} \abs{S \setminus \covDR} \textrm{ for each } \dr \in \ds$\;
        draw $\dr \sim \wgt\pr{\dr}$\;
        draw $\rtail \in \invmap\spr{\dr}  \sim \abs{\rtail}$\;
        \Return{\rtail}
\end{algorithm}

\spara{Constructing \reducedss.}
To construct the interpretable label space, 
we first define interpretability in our setting.
Humans like to think in an associative manner~\cite{morewedge2010associative}. 
To accommodate such tendency, we argue that a label set is interpretable if the corresponding labels are sufficiently associated.
The problem of constructing \reducedss is then framed as finding sufficiently associated label sets.

We rely on a graph-based approach whereby we construct a suitable label graph and extract its dense subgraphs. 
Specifically, we construct a directed weighted graph $\ugdef$. 
Each node represents a label.  %
A node pair $(u,v)$ is an edge in $\edges$ if $\ds\spr{\set{u}} \cap \ds\spr{\set{v}} \neq \emptyset$. 
The corresponding weight is defined as 
$p(u, v) = \frac{\abs{\ds\spr{\set{u}} \cap \ds\spr{\set{v}}}}{\abs{\ds\spr{\set{u}}}}$, 
which can be interpreted as the conditional probability that label $v$ occurs given that label $u$ occurs.
The need of a directed graph arises because in real-world multi-label datasets, 
association of labels is asymmetric.

Finally,
probabilistic interpretation of the edge weights suggests that $\ug$ 
can be viewed as a \textit{probabilistic graph}~\cite{zou2010mining}.
Under such point of view, 
our problem can be seen as finding highly probable cliques in $\ug$~\cite{mukherjee2015mining},
whose probability of forming is above pre-specified threshold.
To solve this problem, 
we adapt a depth-first search (DFS) procedure similar to the one proposed
by Mukherjee et al.~\cite{mukherjee2015mining}.

\spara{Efficient pre\-processing.}
Execution of line~\ref{sampling-red:preproc} in Algorithm~\ref{alg:sampling-red} can be done efficiently by framing the problem appropriately.
In this problem, we are given a set of subsets \reducedss and we are asked to find, for each $\dr \in \ds$, the subsets in \reducedss that are contained in $\labelsD$.
A naive solution checks the containment relations for all pairs of $\labelsD$ and $\reducedss$, and in practice can take hours %
for many datasets.
However, the problem is an instance of the \emph{the set containment problem}, extensively studied by the database community. 
Among several efficient solutions proposed for this problem, we resort to one well-established algorithm, PRETTI~\cite{jampani2005using}, %
built upon the idea of inverted index and prefix trees. The running time is effectively brought down to a few seconds.

\subsection{Improved head sampling}\label{sec:improvements_head}

To alleviate the heavy-hitter problem during head sampling, 
we consider two approaches.
The first approach is based on reduced sample space, but may have scalability issues.
The second is a greedy heuristic, which explicitly maximizes a modified version of discriminativity.

\spara{1.~~Using reduced sample space.}
We adapt a similar idea as in tail sampling (Section~\ref{sec:surs}) 
and use a reduced sample space \reducedss for head sampling.
However, when the feature matrix is dense, a scalability issue arises. 
The DFS procedure may take exponential time. 
For sufficiently sparse graphs this is not a concern in practice, 
whereas in denser graphs, constructing \reducedss becomes a bottleneck.

\spara{2.~~A greedy heuristic.}
To address the above scalability issue, we propose a greedy heuristic, 
which drops the probabilistic guarantee, but is highly effective in practice.
We use \cftp as in Algorithm~\ref{alg:headSampling} to sample a tuple \tupledef.
Then %
we greedily select features in $\posD \setminus \negD$ to maximize a modified version of discriminativity:
for any \rhead, we define the measure
\begin{equation}
  \label{eq:modified-discriminativity}
  \phi(\rhead) = \abs{\ds\spr{\rhead} \cap \positives} - \gamma \abs{\ds\spr{\rhead} \cap \negatives},
\end{equation}
where $\gamma$ weighs the importance of positive and negative support, 
so smaller values of $\gamma$ lead to more general but more error-prone heads. 
Further, we use early stopping (controlled by $\epsilon$) when $\abs{\ds\spr{H}}$ is too small.

The algorithm is described in Algorithm~\ref{alg:greedy_discr}.
It iteratively picks a feature $h \in \feats_{\posD} \cup \feats_{\negD}$, 
which maximizes the marginal gain of $\phi$. %
The best feature $h^*$ is added to $\currH$ and the support is updated accordingly.
Finally, a linear sweep over $\currH$ finds the head with the highest objective value 
(in~\eqref{eq:modified-discriminativity}).
In practice, we use a pre-computed inverted index 
to allow for efficient intersection of supports. Variations of Algorithm~\ref{alg:greedy_discr} have been investigated in which the input is deterministic, the difference in line \ref{line:margi} is normalized by $\ds[\{h\}]$ and the support is replaced by support not covered by previously chosen rules.

\spara{Summary.}
The second approach scales better for dense feature matrices than the first approach. 
However, the first approach has the following advantages:
(1) head sampling has probabilistic guarantees, 
(2) it is much faster to run when the feature matrices are sparse.
In the sequel, 
we use \ouralgsurs to denote the version where the first approach is used for head sampling, 
and \ouralggh when the second approach is used.

\begin{algorithm}[t]
	\SetAlgoLined
	\KwData{\hspace*{-2mm}
		a dataset \ds, sets \positives, \negatives, parameters $\gamma$ and $\epsilon$. 
	}
	\KwResult{a head $\rhead \in \featspace$.}
	\caption{
          A greedy heuristic for head sampling. 
		\label{alg:greedy_discr}
              }
              let $\featspool \leftarrow \feats_{\posD} \setminus \feats_{\negD}$\;
              initialize $\currH \gets \text{an empty list}, \discrscores \gets \text{an empty list}$\;
	\While{$|\currH| < |\featspool|$}{
          $h^* \gets \argmax_{h \in \featspool} \spr{\phi(H \cup \set{h}) - \phi(H)}$\;\label{line:margi} %
          add  $h^*$ to  $\currH$,~ add  $\phi(\currH)$ to  $\discrscores$ \;
          $\featspool \gets \featspool \setminus h^*$\;
          \lIf{$|\ds[\currH]| < \epsilon \;  |\positives|$}
          {break}
	}
        $i^{*} \leftarrow \argmax_{i=1,\ldots,\abs{\currH}} \discrscores[i]$\;
	\Return $\currH[1:i^{*}]$
\end{algorithm}

%% file: complexity.tex
\section{Complexity Analysis}
\label{sec:complexity}

\newcommand{\timecost}{\ensuremath{T}}
\newcommand{\spacecost}{\ensuremath{S}}

\spara{Time complexity.}
Let $\timecost_\emph{f}$ be the time complexity of evaluating the quality and diversity function.
$\timecost_\emph{f}$ is bounded by $|\dataset| ( |\featspace| +  |\labelspace| )$.
Let $\reducedss_{\labelspace}$ be the interpretable sample space for tail sampling and $\reducedss_{\featspace}$  be the reduced sample space for head sampling. 
The pre-processing times $\timecost_\emph{S}^{\labelspace}$ and 
$\timecost_\emph{S}^{\featspace}$ 
to construct $\reducedss_{\labelspace}$ and $\reducedss_{\featspace}$ 
are exponential in the worst case.
It follows that the time complexity of \ouralg is 
$\bigO( B  \abs{\candidaterules}  \timecost_\emph{f} +  \timecost_\emph{S}^{\labelspace} +  \timecost_\emph{S}^{\featspace} )$. 
If \featspace is sufficiently sparse, the exponential complexity is not a concern in practice. When \featspace is dense, it is appropriate to use \ouralgorithmplus, 
for which the time complexity is 
$\bigO( B  \abs{\candidaterules}  \timecost_\emph{f} +  \timecost_\emph{S}^{\labelspace} )$. 

\spara{Space complexity.}
Space 
$\spacecost_\emph{\crule}\!=\!\bigO(|\dataset| + |\featspace| + |\labelspace|)$
is required to keep a single rule,
as we store head, tail, and coverage.
Let $\spacecost_\emph{S}$ denote the space complexity of sampling.
In both tail and (owing to \cftp) head sampling, 
we only need to store a single weight value for each data record.
Building $\reducedss_{\labelspace}$ and $\reducedss_{\featspace}$  
requires space $\bigO(|\featspace|^2)$ and $\bigO(|\labelspace|^2)$, respectively.
Furthermore, storing samples from  $\reducedss_{\labelspace}$ ($\reducedss_{\featspace}$) 
takes space $\bigO(\abs{\dataset}\abs{\reducedss_{\labelspace}})$ 
($\bigO(\abs{\dataset}\abs{\reducedss_{\featspace}})$). 
Despite this theoretical complexity, the graphs are very sparse in practice.
Combining the above, we have that 
$\spacecost_\emph{S}\!=\!\bigO(\abs{\labelspace}^2 + \abs{\featspace}^2 + \abs{\dataset} \abs{\reducedss_{\labelspace}} + \abs{\dataset} \abs{\reducedss_{\featspace}})$  
and the space complexity of \ouralg is
$\bigO( ||\featspace|| + ||\labelspace|| +  \abs{\candidaterules}\spacecost_\emph{\crule} + \spacecost_\emph{S})$.
When \ouralgorithmplus is used, the greedy head sampler only takes space $\bigO(|\featspace|)$ and hence
$\spacecost_\emph{S}$ reduces to 
$\spacecost_{\emph{S}+}\!=\!\bigO(\abs{\labelspace}^2 + \abs{\dataset} \abs{\reducedss_{\labelspace}} + |\featspace|)$ so that
the space complexity of \ouralgorithmplus is 
$\bigO( ||\featspace|| + ||\labelspace|| +  \abs{\candidaterules}\spacecost_\emph{\crule} + \spacecost_{\emph{S}+})$.

%% file: exp.tex
\section{Experimental Evaluation}
\label{sec:exp}

The main goal in this section is to empirically show that \ouralg (and in particular its two implementations \ouralgorithmsurs and \ouralggh)
deliver a concise set of rules while still providing competitive performance in multi-label classification. 
We first present the experimental setup and then the results. 

\subsection{Experimental setup}
\label{sec:experimental_setting}

\spara{Datasets.}
We use both synthetic and real-world datasets. 

We use synthetic datasets to better understand the behavior of the methods with respect to different parameters.
Data are obtained from a set of generating rules, 
and as a consequence, a notion of ground truth is available. 
For each generating rule, we sample its support either 
($i$) uniformly at random, or  
($ii$) from a skewed distribution 
where a small subset of rules covers a large portion of the data, mimicking the typical behaviour of real-world data. 
All generating rules have the same number of attributes and labels.
Thus, to obtain the synthetic dataset, we start with a feature and label matrices in which all entries are identically $0$. 
Then, for each rule, once its support is sampled, we set to $1$ its attributes and labels over its support. 
	
For real-world data, we use heterogeneous benchmark datasets for multi-label classifications%
\footnote{http://mulan.sourceforge.net, https://www.uco.es/kdis/mllresources/}.
Summary statistics of the datasets are shown in Table~\ref{tab:datasets}. 
Categorical and numerical features are converted to binary form. 
For simplicity, we convert numerical features into binary ones 
by setting to $0$ all values lower than a given percentile $\mathit{p}$ 
($90$-th percentile by default) and by setting to $1$ the rest of the values. 
A more refined pre-processing is advisable to improve performance.

\spara{Metrics.}
To measure the quality of a classifier, 
we use the popular balanced $F_1$ score, 
which micro-averages precision and recall. 
To monitor rule diversity, 
we report the average pairwise intersection between the coverage of different rules. 
To assess interpretability 
we report the number of rules $\ruleset$. 
\begin{table}
\begin{center}
		\caption{\label{tab:datasets} Summary statistics of the datasets used in the experimental evaluation. The last two columns refer to the average number of labels per example, and the total number of distinct label sets.}
		\footnotesize
		\begin{tabular}{l r r r r r}
			\toprule
			Dataset & Instances  & Attributes & Labels & Cardinality & Distinct \\  
			\midrule
			\mediamillf & 43\,907 & 120 & 101 & 4.38 & 6\,555\\ 
			\Yelpf & 10\,810 & 671 & 5 & 1.64 & 32\\
			\corelf & 5\,000 & 499 & 374 & 3.52 & 3\,175 \\
			\bibtexf  & 7\,395 & 1\,836 & 159 & 2.40 & 2\,856\\ 
			\enronf & 1\,702 & 1\,001 & 53 & 3.38 & 753 \\ 
			\medicalf & 978 & 1\,449 & 45 & 1.24 & 94 \\ 
			\birdsf & 645 &  260 & 19 & 1.01 & 133\\ 
			\emotionsf & 593 & 72 & 6 & 1.87 &27 \\
			\CALf & 502 & 68 & 174 & 26.04 & 502 \\ 
			\bottomrule 
		\end{tabular}

\end{center}
\end{table}

\spara{Baselines.}
We compare our classifier with three baselines. 

\smallskip
\noindent
\seco~\cite{klein2019efficient} is a rule-based classifier, which extracts new rules
iteratively and discards the associated covered examples from the training data if enough of their labels are predicted by already learned rules. 
Given a rule head, 
\seco searches for the best possible tail according to a metric, 
while pruning the search space by exploiting properties of the metric, and introducing bias towards tails with multiple labels. 

\smallskip
\noindent	
\Boomer~\cite{rapp2020learning, boley2021better} utilizes the gradient-boosting framework 
to learn ensembles of single-label or multi-label classification rules 
that are combined additively to minimize the
empirical risk with respect to a suitable loss function. 

\smallskip
\noindent
\SVM~\cite{zhang2018binary, cortes1995support} is a linear support vector machine classifier 
based on the binary relevance approach, whereby each label is treated independently. 
This classifier is not rule-based, and serves as a black-box baseline. 
\smallskip

In general, \Boomer 
takes advantage of a large number of rules, 
which are then combined to generate the final scores 
from which the predictions are derived. 
In this way, it achieves state-of-the-art performance 
in associative multi-label classification. 
In addition, it controls the number of rules
in the ensemble with a single parameter. 
Thus, it is the most important baseline. 
For the synthetic datasets, we focus on comparing our approach with \Boomer 
for increasing number of rules, 
whereas for the real-world datasets we consider all baselines. 

\spara{Parameter setting.}
For the experiments with synthetic data, 
we explore the scalability of our algorithm 
with respect to the number of attributes and labels, 
as well as robustness with respect to noise. 
We vary the level of noise 
(proportion of flipped entries in the feature and label matrix), 
and  the number of attributes and labels 
by a geometric progression of ratio $1.5$. 
When not varied, the number of attributes and labels are fixed to $100$, 
and the noise level to $0.01$. 
When the noise is varied there are $10$ ground truth rules, 
otherwise the number of generating rules increases 
with the size of the data and it is given by 
$ \lfloor  \frac{\min(  |\mathcal{F}|, |\mathcal{L}| ) }{ 3 } \rfloor $. 

For the experiments with real-world data, 
we tune the hyper-parameters of all methods 
via random search to minimize micro-averaged $F_1$ on a validation set. 
The size of each sampled pool of rules $\currentrules$ does not need to be tuned. 
Larger $\currentrules$ improves performance at the cost of increased runtime. 
While tuning we fix $\currentrules=150$, 
otherwise $\currentrules$ is set to $500$ by default. 
All hyper-parameters are searched in the range $(0,1)$, except for $\lambda$ that is searched in $(10^{-2}, 10^{2})$.  
We also investigate the impact of $\lambda$ on the the diversity 
of the set of chosen rules \ruleset, 
by varying it in a geometric progression of ratio $10$. 
All experiments results are obtained as average 
over $10$ repetitions to account for randomness.

\spara{Implementation.} Experiments are executed on a machine with
$2\!\times\!10$\,core Xeon\,E5\,2680\,v2\,2.80\,GHz processor and 256\,GB memory. 
Our implementation is available online%
\footnote{https://github.com/DiverseMultiLabelClassificationRules/CORSET}. 
To speed-up and facilitate hyper-parameter tuning for \ouralgorithm we have implemented two practical changes. 
First, we run only the first round of greedy selection in Algorithm~\ref{alg:CMRS}.
The second round guarantees the approximation factor, 
but often offers a modest increase in performance, 
not worth the increase in running time. 
Second, we pass as input to \ouralgorithm the number of rules to be returned (at most $150$)
instead of the tolerance parameter ($c$ in Algorithm~\ref{alg:CMRS}) 
to reduce variability and simplify hyperparameter optimization. 

\subsection{Results}\label{sec:results}

\begin{table*}
	\caption{\label{tab:f1scoresBoomer} 
		Micro-averaged $F_1$-scores on real datasets achieved by 
		\ouralgsurs, \ouralggh, \seco, \Boomer 
		with increasing number of rules, and \SVM. 
		The last three columns show the number of rules for 
		\ouralgsurs, \ouralggh, and \seco.}
	\centering
	\small
	\begin{tabular}{l r r r r r r r r r r}
		\toprule
		Dataset &  \ouralgsurs & \ouralggh & \seco & Boomer & Boomer  & Boomer & BR-SVM & $|\ruleset|$ & $|\ruleset|$ & $|\ruleset|$ \\
		&  &  &  & (10) & (100) & (1000) &  & \ouralgorithmsurs & \ouralgorithmplus & \seco \\
		\midrule
		\mediamillf & 0.44 & 0.51 & NA & 0.43 & 0.44 & 0.50 & 0.50 & 150 & 150 & NA \\ 
		\Yelpf & 0.66 & 0.64 & NA & 0.47  & 0.63 & 0.75 & 0.70  & 67 & 82 & NA \\ 
		\corelf & 0.18 & 0.18 & NA & 0.00 & 0.00 & 0.03 & 0.16 & 142 & 150 & NA \\ 
		\bibtexf & 0.36 & 0.40 &  NA & 0.00 & 0.13 & 0.36 & 0.41 & 74 & 150 & NA \\ 
		\enronf & 0.55 & 0.53 & NA & 0.39 & 0.47 & 0.54 & 0.52 & 41 & 48 &  NA \\ 
		\medicalf & 0.81 & 0.83 & 0.63 & 0.00 & 0.50 & 0.91 & 0.99 & 27 & 88 & 199 \\ 
		\birdsf & 0.37 & 0.42 & 0.39 &  0.00 & 0.34 & 0.46 & 0.42 & 42 & 48 &  122 \\ 
		\emotionsf & 0.53 & 0.54 & 0.53 & 0.17 & 0.49 & 0.54 & 0.56 & 42 & 68 & 199 \\ 
		\CALf & 0.29 & 0.32 & NA & 0.31 & 0.31 & 0.33 & 0.53 &  150 & 150 & NA \\ 
		\bottomrule  
	\end{tabular}
\end{table*}

\begin{figure}
	\centering
	\includegraphics[width=1\columnwidth, height = 0.16\textheight, keepaspectratio]{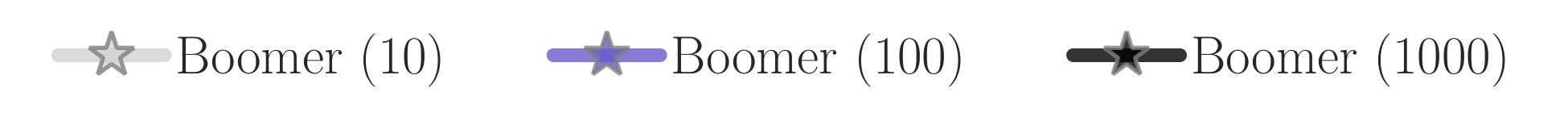}\\
	\includegraphics[width=1\columnwidth, height = 0.16\textheight, keepaspectratio]{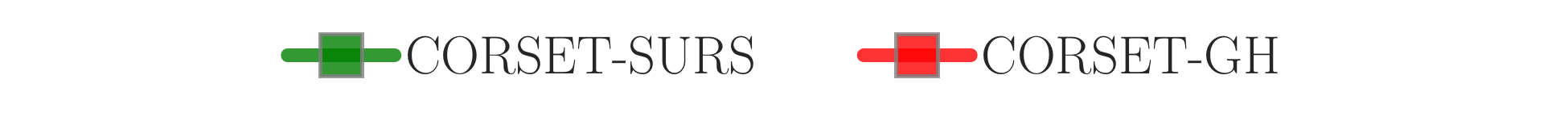}
\begin{tabular}{ccc}
		\includegraphics[width=0.3\columnwidth, height = 0.25\textheight, keepaspectratio]{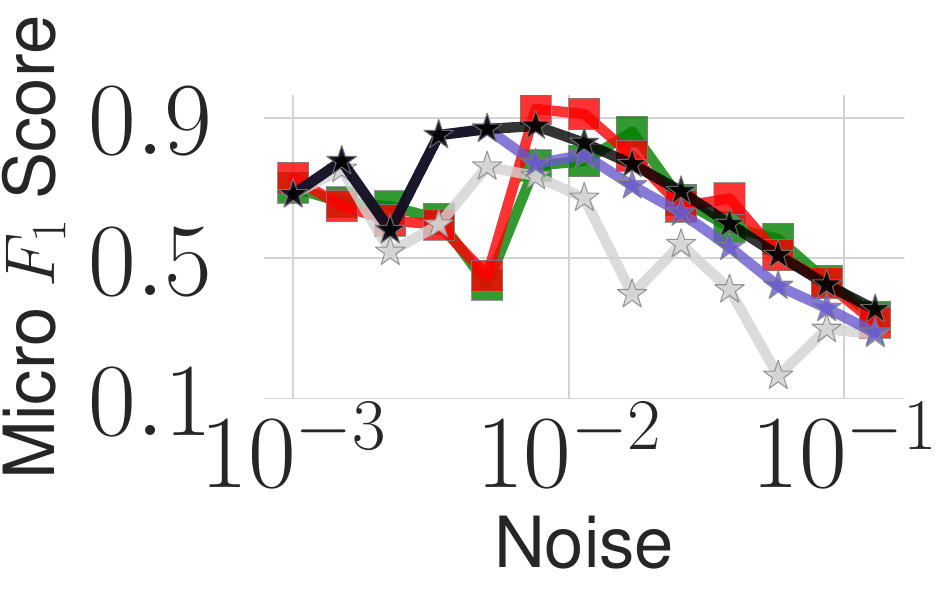}&
		\includegraphics[width=0.3\columnwidth, height = 0.25\textheight, keepaspectratio]{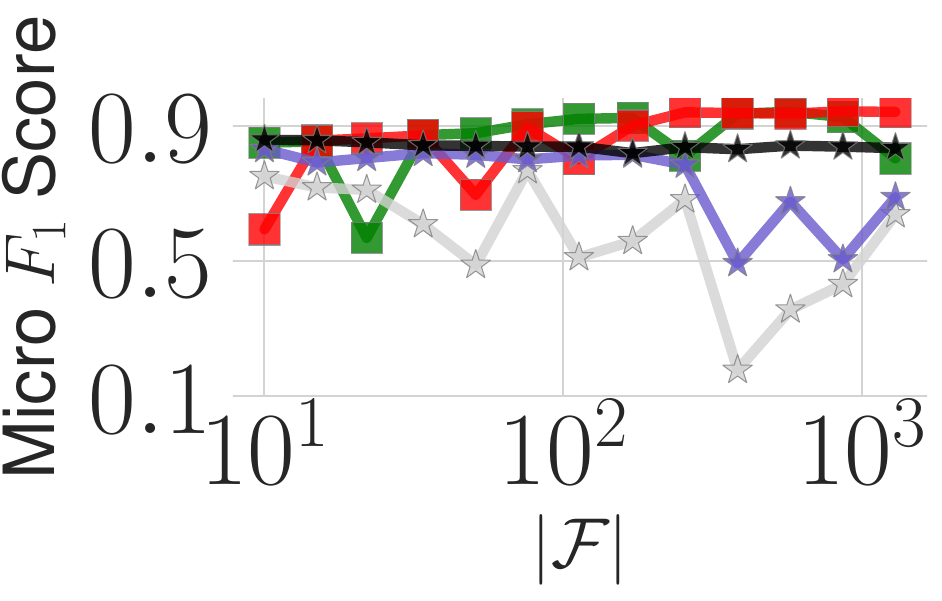}& 
		\includegraphics[width=0.3\columnwidth, height = 0.25\textheight, keepaspectratio]{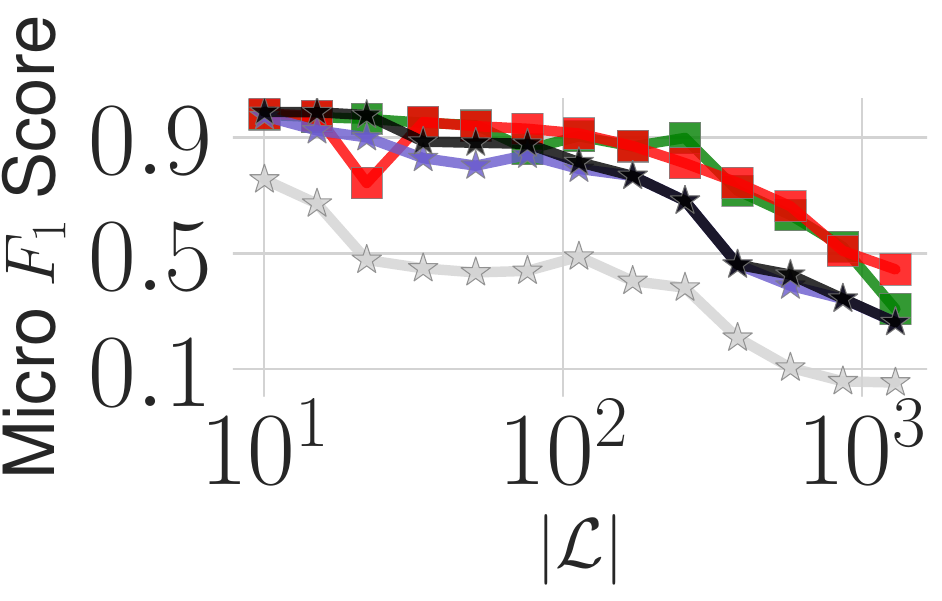}\\
		\includegraphics[width=0.3\columnwidth, height = 0.25\textheight, keepaspectratio]{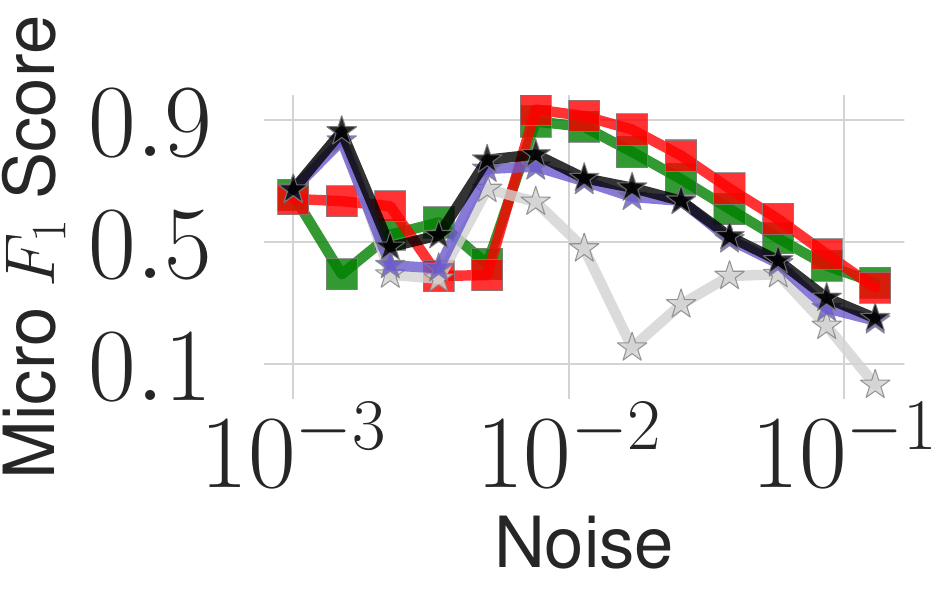}&
		\includegraphics[width=0.3\columnwidth, height = 0.25\textheight, keepaspectratio]{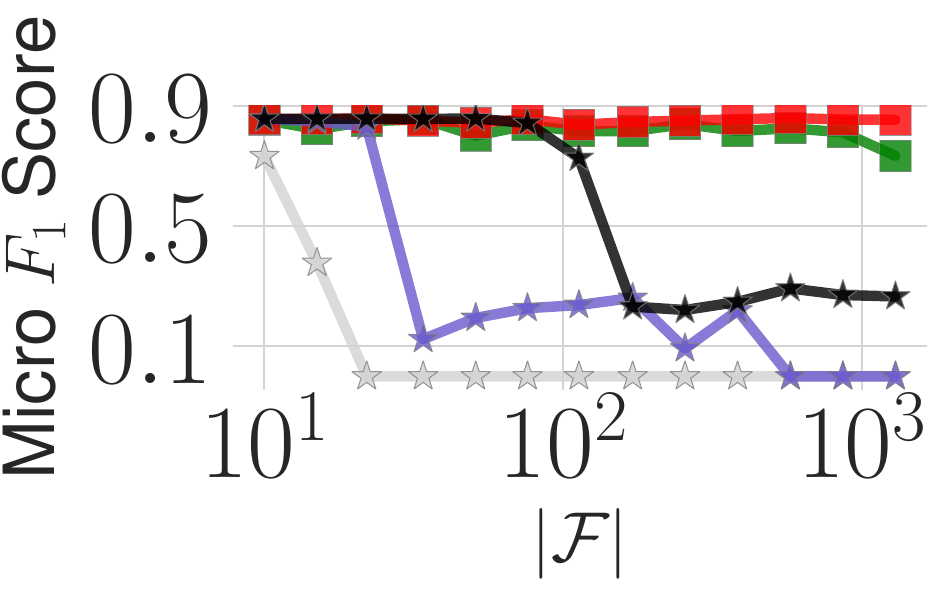}& 
		\includegraphics[width=0.3\columnwidth, height = 0.25\textheight, keepaspectratio]{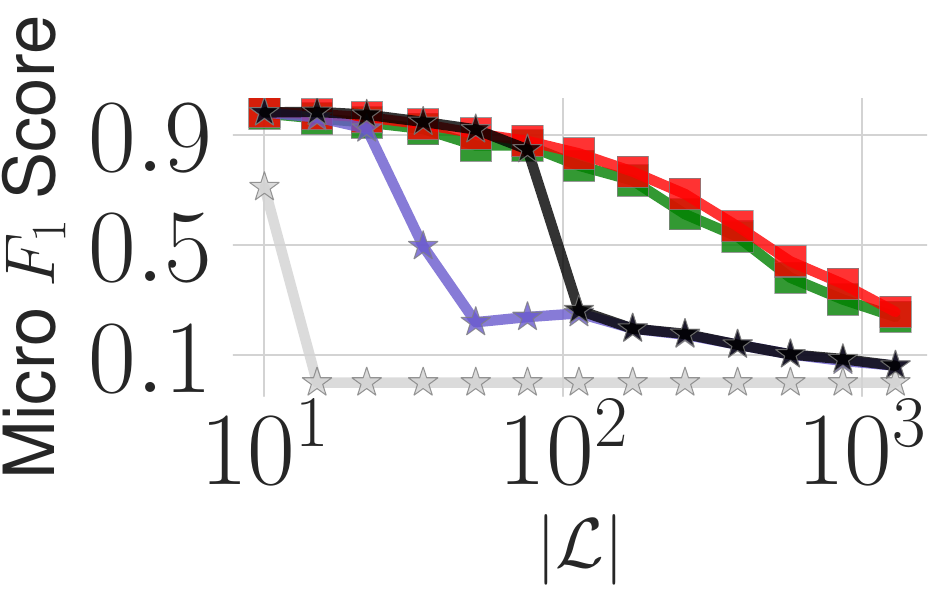}\\
	\end{tabular}
	\vspace{-0.2cm}
	\caption{\label{fig:synthetic_datasets} 
	Synthetic datasets generated from rules with uniform (top) and 
	skewed (bottom) coverage. 
	Micro-averaged $F_1$ score against proportion of noise (left), 
	number of attributes (middle) and labels (right). 
	The $x$-axis is in log scale. } 
\end{figure}

\spara{Synthetic datasets.}
Results on synthetic datasets, both for 
data generated from rules with uniform and skewed coverage, 
are shown in Fig.~\ref{fig:synthetic_datasets}.
The number of rules retrieved by \ouralgorithm 
is at most the number of generating rules, $33$. 
On the other hand, \Boomer based on $10$ rules consistently offers poor performance. 
The classification accuracy of \Boomer increases when the number of rules increases, 
but even with $1000$ rules, 
our method outperforms \Boomer while using a very concise set of rules. 
Unlike \Boomer, \ouralgorithm seeks to uncover the true set of generating rules 
and only use those for classification. 
Thus, the experiments with synthetic datasets clearly show the advantage of our approach. 
Also note that the performance of \ouralgorithm, unlike that of \Boomer, 
does not significantly deteriorate when $|\featspace|$ increases.

\spara{Real datasets: classification performance and interpretability.}
Results on real datasets, for classification performance and interpretability, 
are shown in Table~\ref{tab:f1scoresBoomer}.
In some cases, \seco does not terminate within a time interval of $12$ hours, 
in such cases we report NA in the corresponding table entry.
Table~\ref{tab:f1scoresBoomer} shows that \Boomer requires a very large number of rules to achieve competitive performance. 
Thus, it does not offer high interpretability. 
Similarly, \SVM performs well but it is not interpretable. 
Instead, \ouralgorithm extracts a small set of rules, 
guaranteeing ease of interpretation, 
and yet it is consistently competitive with the baselines 
on all the datasets. 
\ouralgorithm always requires fewer rules than rule-based 
alternatives to attain the same performance in multi-label classification, and 
it is never drastically worse than \Boomer with $1000$ rules or even \SVM, suggesting that the \emph{price of interpretability}, 
if there is one, is small when \ouralgorithm is used. 
Finally, note that \ouralggh often outperforms \ouralgorithmsurs but using a larger set of rules. 

\begin{figure}
	\centering
	\hspace*{-0.5cm}\includegraphics[width=0.6\columnwidth, height = 0.4\textheight, keepaspectratio]{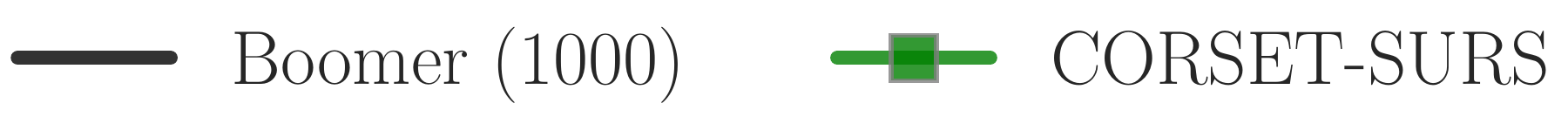}
\begin{tabular}{ccc}
	\hspace*{-0.16cm}\includegraphics[width=0.32\columnwidth, height = 0.4\textheight, keepaspectratio]{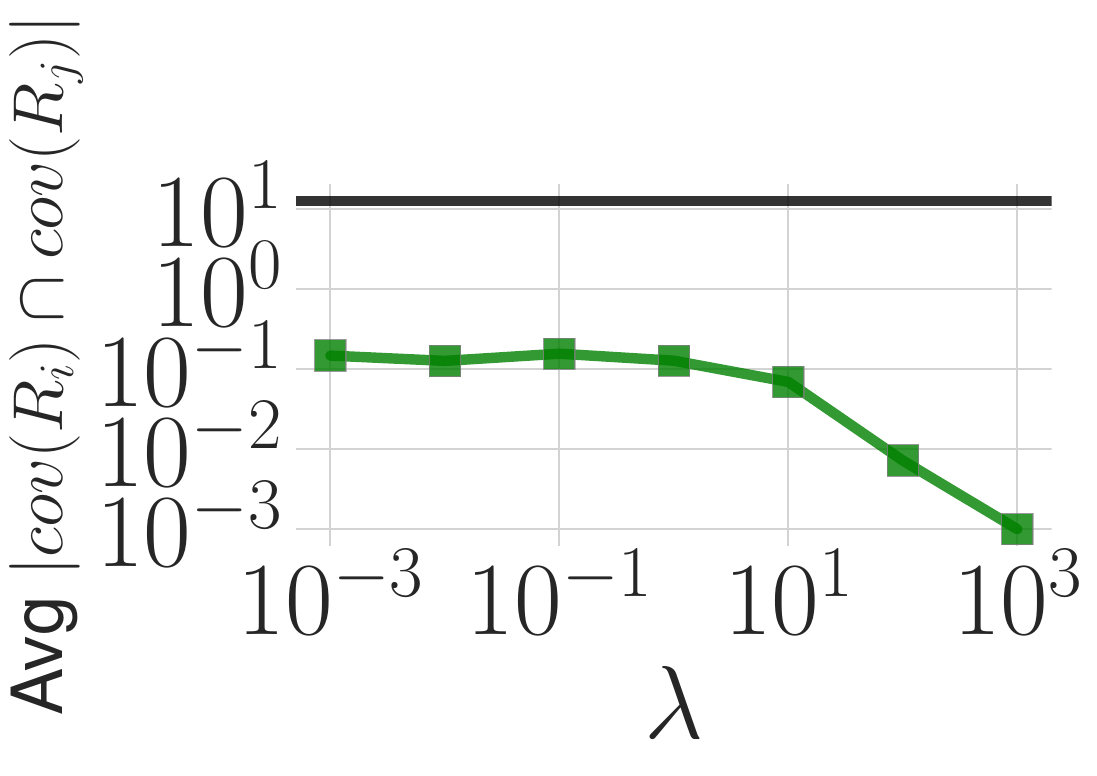}&
	\hspace*{-0.16cm}\includegraphics[width=0.32\columnwidth, height = 0.4\textheight, keepaspectratio]{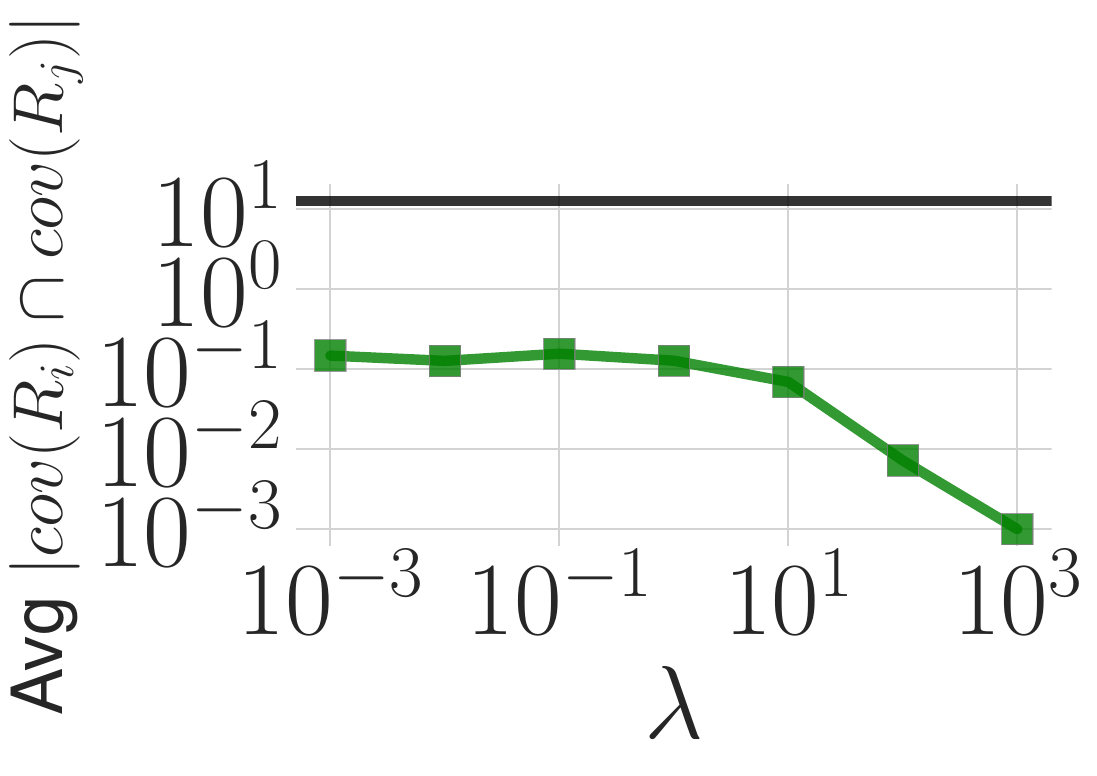}& 
	\hspace*{-0.16cm}\includegraphics[width=0.32\columnwidth, height = 0.4\textheight, keepaspectratio]{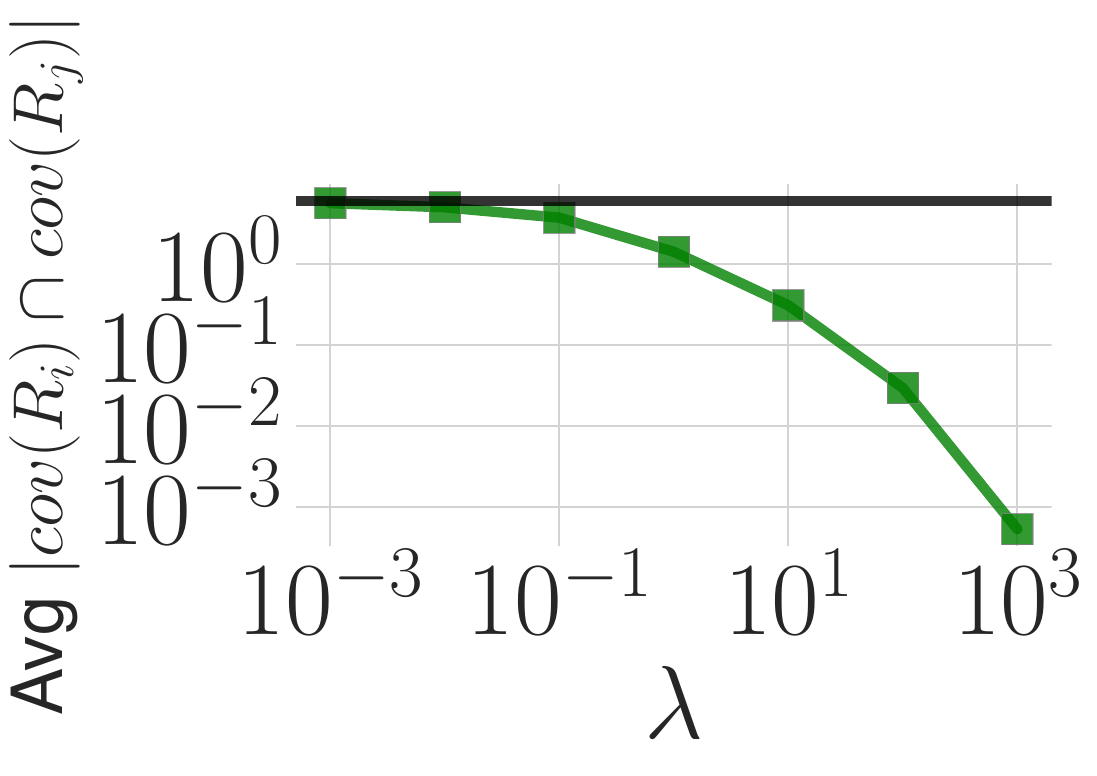}\\
    \hspace*{-0.16cm}(a) \corel & (b) \bibtex  &  (c)  \enron \\ 
	\hspace*{-0.16cm}\includegraphics[width=0.32\columnwidth, height = 0.4\textheight, keepaspectratio]{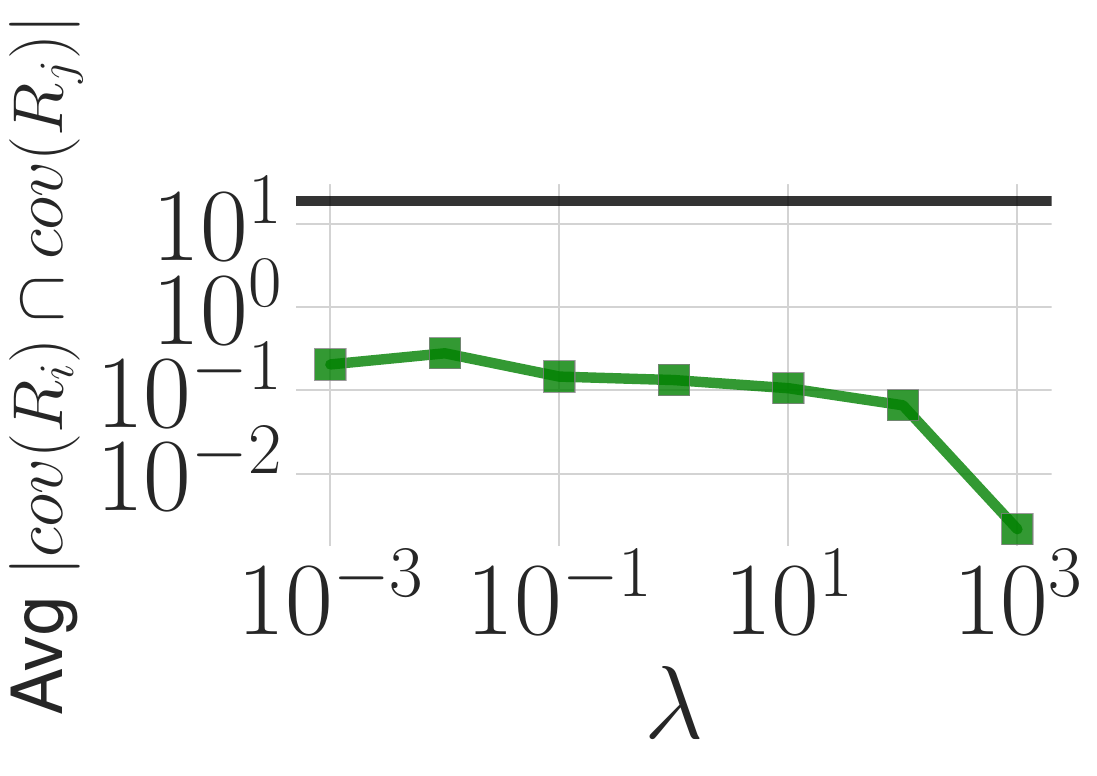}&
	\hspace*{-0.16cm}\includegraphics[width=0.32\columnwidth, height = 0.4\textheight, keepaspectratio]{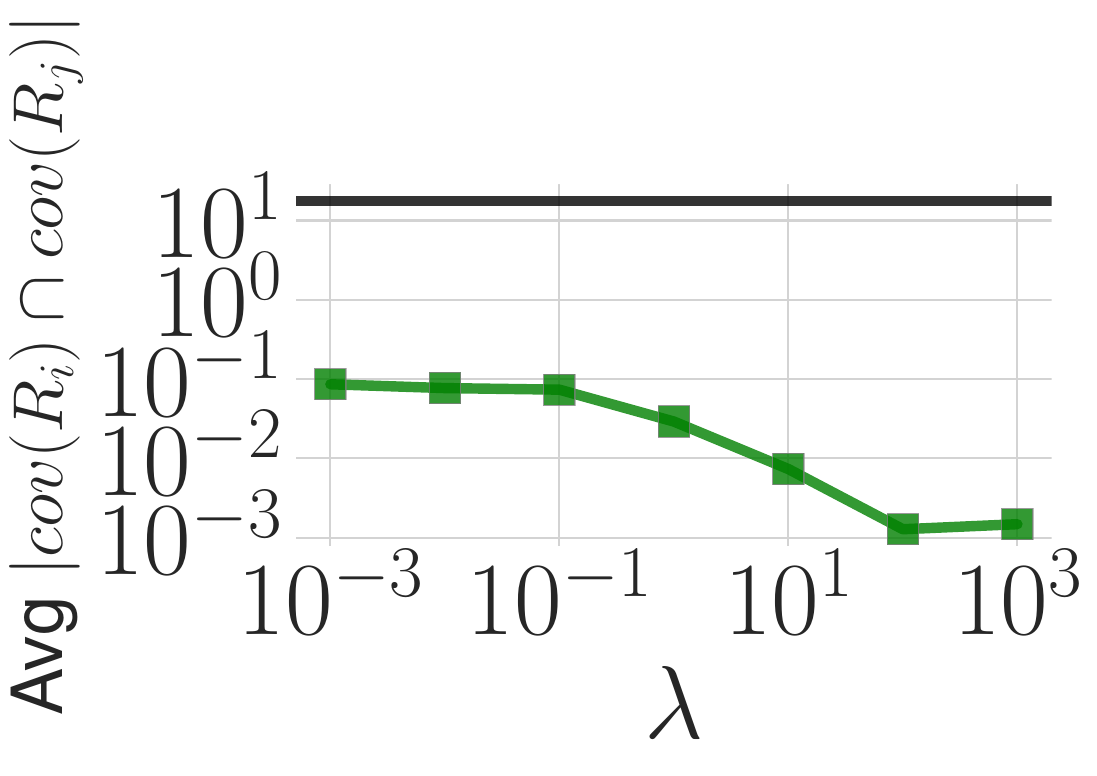}& 
	\hspace*{-0.16cm}\includegraphics[width=0.32\columnwidth, height = 0.4\textheight, keepaspectratio]{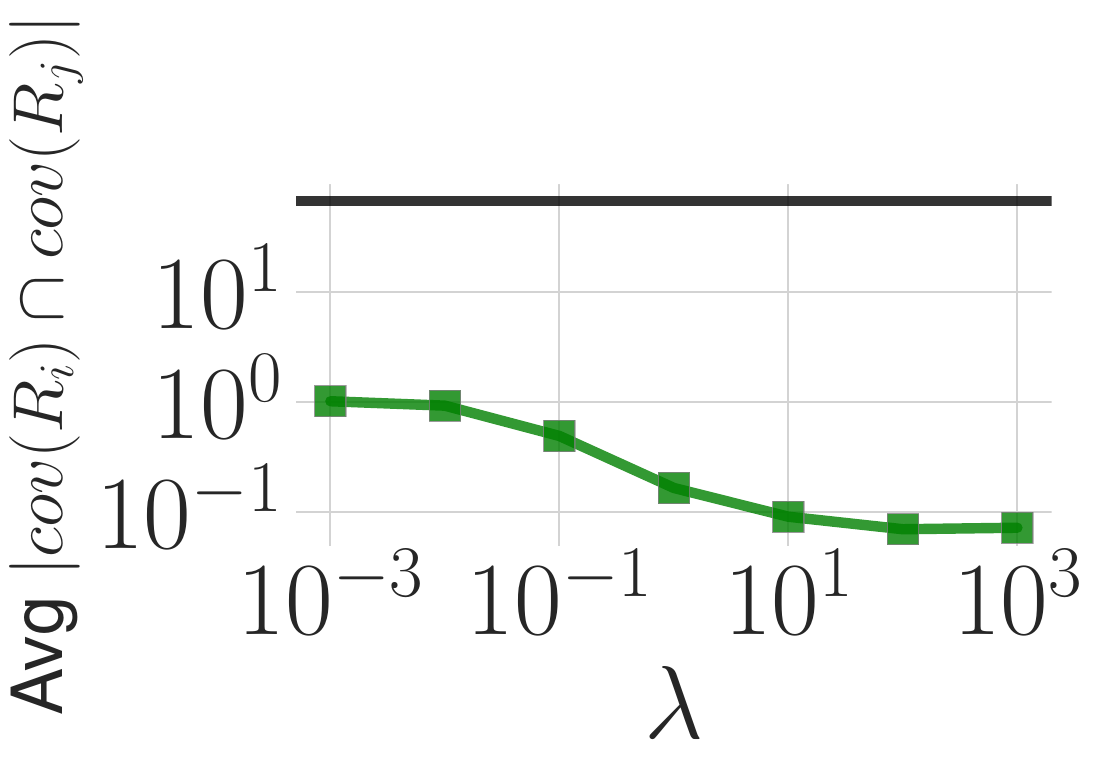}\\
	\hspace*{-0.16cm}(d) \medical & (e) \birds &  (f)  \emotions
\end{tabular}
	\caption{\label{fig:lambda_diversity} 
	Real datasets. Average coverage overlap between pairs of rules as a function of $\lambda$ (lower values indicate higher diversity). 
	Both axes are in log scale.} 
\end{figure}

\spara{Real datasets: diversity and impact of $\lambda$.}
A fundamental characteristic of \ouralgorithm
is that it allows to control the degree of diversity
in the set of recovered rules via a single tunable parameter $\lambda$. 
In Fig.~\ref{fig:lambda_diversity}, 
we show for a subset of datasets that the shared coverage within rules is lower 
for \ouralgorithm than for \Boomer, 
and moreover that increasing the value of $\lambda$ is very effective in reducing overlap between rules. In practice $\lambda$ must be carefully tuned to optimize the performance of \ouralgorithm. 
As the impact of $\lambda$ is not significantly different in \ouralgsurs and \ouralggh, we only show results for the former.

%% file: conclusion.tex
\section{Conclusion}
\label{sec:conclusion}

We propose a novel rule-based classifier, \ouralg, for multi-label classification tasks.
Our training objective explicitly penalizes rule redundancy, 
encouraging the algorithm to learn a concise set of rules.
Furthermore, we design a suite of fast sampling algorithms, 
which can generate rules with good accuracy and interpretability.
We show that \ouralg achieves competitive performance 
comparable to strong baselines, while offering better interpretability.

Our work opens interesting questions for future research.
Can we design training objectives that reflect popular multi-label classification metrics, 
while producing concise rule sets?
Can we use the techniques in this work 
to address the interpretability issue of existing rule-based classifiers?

%% file: appendix.tex
\appendix
\section*{Proof of Proposition 1}

To show that \ouralg is a $2$-approximation algorithm for Problem~\ref{pr:main}, we need to show that in the objective function: 

\begin{align*}
	\obj\pr{\ruleset} = \sum\limits_{\crule \in \ruleset} \quality\pr{\crule; \ruleset \setminus \set{\crule}} + \lambda \sum\limits_{\crule_i, \crule_j \in \ruleset, i \neq j} \dist(\crule_i, \crule_j)
\end{align*}

the first term (quality function) is monotone and submodular and the distance junction, namely the Jaccard distance, is a metric.

First, the quality function is a sum of non-negative terms, and hence it is monotone non-decreasing. 
It is straightforward to prove that the function is additionally submodular. Consider two rule sets $\ruleset'$ and $\ruleset''$ such that $\ruleset' \subseteq \ruleset''$. 

Given a new rule $\crule^*$  marginal gain in $\quality\pr{\crule^*; \ruleset'}$ is:
\begin{align*}
\Delta' = \disc\pr{\crule^*} \times \abs{\cov\pr{\crule^*} \setminus \bigcup\limits_{\crule \in \ruleset'} \cov\pr{\crule}}. 
\end{align*}
Similarly, the marginal gain in $\quality\pr{\crule^*; \ruleset''}$ is:
\begin{align*}
  \Delta'' &\; = \disc\pr{\crule^*} \times \abs{\cov\pr{\crule^*} \setminus \bigcup\limits_{\crule \in \ruleset'} \cov\pr{\crule}} \\
  & \;= \disc\pr{\crule^*} \times \abs{\cov\pr{\crule} \setminus \bigcup\limits_{\crule \in \ruleset'} \cov\pr{\crule} \setminus \bigcup\limits_{\crule \in \ruleset'' \setminus  \ruleset'} \cov\pr{\crule}  }.
\end{align*}
Since
\begin{align*}
  & \; \abs{\cov\pr{\crule^*} \setminus \bigcup\limits_{\crule \in \ruleset'} \cov\pr{\crule}} \\
  \geq &\; \abs{\cov\pr{\crule} \setminus \bigcup\limits_{\crule \in \ruleset'} \cov\pr{\crule} \setminus \bigcup\limits_{\crule \in \ruleset'' \setminus  \ruleset'} \cov\pr{\crule}  },
\end{align*} 
it immediately follows that $\Delta'\geq \Delta''$. 
Therefore, we conclude that the quality function is submodular.

As concerns the Jaccard distance, first notice that:
\begin{align*}
\dist\pr{\crule, \crule} = 1 - \frac{\abs{\cov\pr{\crule} \cap \cov\pr{\crule}}}{\abs{\cov\pr{\crule} \cup \cov\pr{\crule}}} = 0. 
\end{align*}
Furthermore, the Jaccard distance is symmetric: 
\begin{align*}
\dist\pr{\crule_1, \crule_2} = \dist\pr{\crule_2, \crule_1} =  1 - \frac{\abs{\cov\pr{\crule_1} \cap \cov\pr{\crule_2}}}{\abs{\cov\pr{\crule_1} \cup \cov\pr{\crule_2}}} \\ =  1 - \frac{\abs{\cov\pr{\crule_2} \cap \cov\pr{\crule_1}}}{\abs{\cov\pr{\crule_2} \cup \cov\pr{\crule_1}}}. 
\end{align*}
Finally,to prove that the Jaccard distance is a metric, it is left to prove that it satisfies the triangle inequality. Several proofs that the triangle inequality holds for the Jaccard distance exist~\cite{kosub2019note}.

Borodin et al.~\cite{borodin2012max} show that the properties we have proved for the quality and distance function in $\obj\pr{\ruleset}$ guarantee that \ouralg is a $2$-approximation algorithm for Problem~\ref{pr:main}.

\section*{Proof of Proposition 2}

We prove that Algorithm~\ref{alg:tailSampling} returns $\rtail \sim \relarea{\rtail}$.

\begin{align*}
& Pr(\rtail \text{ is drawn})=\sum_{\dr \in \ds} Pr(\rtail \text{ is drawn and } \dr \text{ is drawn} )\\
&= \sum_{\dr \in \ds} Pr(\dr \text{ is drawn} ) Pr(\rtail \text{ is drawn from } \powerset\pr{\labelsD})   \\ 
&= \sum_{\dr \in \dsT} Pr(\dr \text{ is drawn} ) Pr(\rtail \text{ is drawn from } \powerset\pr{\labelsD})   \\ 
& \propto \sum_{\dr \in \dsT}  w(\dr;\ruleset) \times \frac{w(\rtail, \dr; \ruleset)}{w(\dr;\ruleset)} \\
&=  \sum_{\dr \in \dsT}w(\rtail, \dr; \ruleset) = \sum_{\dr \in \dsT} \abs{\relcov{\rtail}} = \relarea{\rtail}.
\end{align*}

The first and second equalities follow from the law of total probability, and the chain rule of probabilities, respectively. 
The third equality is guaranteed because $\rtail$ can only be sampled from $\dr \in \dsT$. If $\dr$ is not in $\dsT$, it has null probability of generating \rtail. 
In the fourth equality, we have used:
\begin{equation*}
\sum_{\rtail \subseteq \labelsD} \wgtR{\rtail, \dr} = \wgtR{\dr} . 
\end{equation*}
Finally, the last equality follows since $\relarea{\rtail}$ can be obtained by summing the marginal coverage $\abs{\relcov{\rtail}}$ of $\rtail$ on \dr, given \ruleset, over all data records $\dr \in \dsT$.

%% file: main.bbl
\begin{thebibliography}{10}

\bibitem{boley2011direct}
M.~Boley, C.~Lucchese, D.~Paurat, and T.~G{\"a}rtner.
\newblock Direct local pattern sampling by efficient two-step random
  procedures.
\newblock In {\em Proceedings of the 17th ACM SIGKDD International Conference
  on Knowledge Discovery and Data Mining}, pages 582--590, 2011.

\bibitem{boley2012linear}
M.~Boley, S.~Moens, and T.~G{\"a}rtner.
\newblock Linear space direct pattern sampling using coupling from the past.
\newblock In {\em Proceedings of the 18th ACM SIGKDD International Conference
  on Knowledge Discovery and Data Mining}, pages 69--77, 2012.

\bibitem{boley2021better}
M.~Boley, S.~Teshuva, P.~L. Bodic, and G.~I. Webb.
\newblock Better short than greedy: Interpretable models through optimal rule
  boosting.
\newblock In {\em Proceedings of the 2021 SIAM International Conference on Data
  Mining (SDM)}, pages 351--359, 2021.

\bibitem{borodin2012max}
A.~Borodin, H.~C. Lee, and Y.~Ye.
\newblock Max-sum diversification, monotone submodular functions and dynamic
  updates.
\newblock In {\em Proceedings of the 31st ACM SIGMOD Symposium on Principles of
  Database Systems}, pages 155--166, 2012.

\bibitem{cortes1995support}
C.~Cortes and V.~Vapnik.
\newblock Support-vector networks.
\newblock {\em Machine learning}, 20(3):273--297, 1995.

\bibitem{crammer2003family}
K.~Crammer and Y.~Singer.
\newblock A family of additive online algorithms for category ranking.
\newblock {\em Journal of Machine Learning Research}, 3:1025--1058, 2003.

\bibitem{elisseeff2001kernel}
A.~Elisseeff and J.~Weston.
\newblock A kernel method for multi-labelled classification.
\newblock {\em Advances in neural information processing systems}, 14:681--687,
  2001.

\bibitem{fischer2019sets}
J.~Fischer and J.~Vreeken.
\newblock Sets of robust rules, and how to find them.
\newblock In {\em Joint European Conference on Machine Learning and Knowledge
  Discovery in Databases}, pages 38--54. Springer, 2019.

\bibitem{furnkranz2012foundations}
J.~F{\"u}rnkranz, D.~Gamberger, and N.~Lavra{\v{c}}.
\newblock {\em Foundations of rule learning}.
\newblock Springer Science \& Business Media, 2012.

\bibitem{ghosh2022efficient}
B.~Ghosh, D.~Malioutov, and K.~S. Meel.
\newblock Efficient learning of interpretable classification rules.
\newblock {\em arXiv preprint arXiv:2205.06936}, 2022.

\bibitem{jampani2005using}
R.~Jampani and V.~Pudi.
\newblock Using prefix-trees for efficiently computing set joins.
\newblock In {\em International Conference on Database Systems for Advanced
  Applications}, pages 761--772, 2005.

\bibitem{yk:Relaxed-Pruning}
Y.~Klein, M.~Rapp, and E.~Loza~Menc{\'{\i}}a.
\newblock Efficient discovery of expressive multi-label rules using relaxed
  pruning.
\newblock In P.~Kralj~Novak, T.~{\v S}muc, and S.~D{\v z}eroski, editors, {\em
  Discovery Science}, pages 367--382. Springer International Publishing, Oct.
  2019.
\newblock Best Student Paper Award.

\bibitem{klein2019efficient}
Y.~Klein, M.~Rapp, and E.~Loza~Menc{\'\i}a.
\newblock Efficient discovery of expressive multi-label rules using relaxed
  pruning.
\newblock In {\em International Conference on Discovery Science}, pages
  367--382. Springer, 2019.

\bibitem{kosub2019note}
S.~Kosub.
\newblock A note on the triangle inequality for the jaccard distance.
\newblock {\em Pattern Recognition Letters}, 120:36--38, 2019.

\bibitem{liu1998integrating}
B.~Liu, W.~Hsu, Y.~Ma, et~al.
\newblock Integrating classification and association rule mining.
\newblock In {\em Kdd}, volume~98, pages 80--86, 1998.

\bibitem{miller2019explanation}
T.~Miller.
\newblock Explanation in artificial intelligence: Insights from the social
  sciences.
\newblock {\em Artificial intelligence}, 267:1--38, 2019.

\bibitem{morewedge2010associative}
C.~K. Morewedge and D.~Kahneman.
\newblock Associative processes in intuitive judgment.
\newblock {\em Trends in cognitive sciences}, 14(10):435--440, 2010.

\bibitem{mukherjee2015mining}
A.~P. Mukherjee, P.~Xu, and S.~Tirthapura.
\newblock Mining maximal cliques from an uncertain graph.
\newblock In {\em 2015 IEEE 31st International Conference on Data Engineering},
  pages 243--254. IEEE, 2015.

\bibitem{rapp2021gradient}
M.~Rapp, E.~L. Menc{\'\i}a, J.~F{\"u}rnkranz, and E.~H{\"u}llermeier.
\newblock Gradient-based label binning in multi-label classification.
\newblock {\em arXiv preprint arXiv:2106.11690}, 2021.

\bibitem{rapp2020learning}
M.~Rapp, E.~L. Menc{\'\i}a, J.~F{\"u}rnkranz, V.-L. Nguyen, and
  E.~H{\"u}llermeier.
\newblock Learning gradient boosted multi-label classification rules.
\newblock {\em arXiv preprint arXiv:2006.13346}, 2020.

\bibitem{read2008pruned}
J.~Read.
\newblock A pruned problem transformation method for multi-label
  classification.
\newblock In {\em New Zealand Computer Science Research Student Conference},
  page~41, 2008.

\bibitem{thabtah2005mcar}
F.~Thabtah, P.~Cowling, and Y.~Peng.
\newblock Mcar: multi-class classification based on association rule.
\newblock In {\em The 3rd ACS/IEEE International Conference onComputer Systems
  and Applications, 2005.}, page~33, 2005.

\bibitem{thabtah2004mmac}
F.~A. Thabtah, P.~Cowling, and Y.~Peng.
\newblock Mmac: A new multi-class, multi-label associative classification
  approach.
\newblock In {\em Fourth IEEE International Conference on Data Mining
  (ICDM'04)}, pages 217--224, 2004.

\bibitem{tidake2018multi}
V.~Tidake and S.~Sane.
\newblock Multi-label classification: a survey.
\newblock {\em International Journal of Engineering and Technology},
  7(4.19):1045--1054, 2018.

\bibitem{wang2011approach}
X.~Wang, K.~Yue, W.~Niu, and Z.~Shi.
\newblock An approach for adaptive associative classification.
\newblock {\em Expert Systems with Applications}, 38(9):11873--11883, 2011.

\bibitem{zhang2020diverse}
G.~Zhang and A.~Gionis.
\newblock Diverse rule sets.
\newblock In {\em Proceedings of the 26th ACM SIGKDD International Conference
  on Knowledge Discovery and Data Mining}, pages 1532--1541, 2020.

\bibitem{zhang2018binary}
M.-L. Zhang, Y.-K. Li, X.-Y. Liu, and X.~Geng.
\newblock Binary relevance for multi-label learning: an overview.
\newblock {\em Frontiers of Computer Science}, 12(2):191--202, 2018.

\bibitem{zhang2007ml}
M.-L. Zhang and Z.-H. Zhou.
\newblock Ml-knn: A lazy learning approach to multi-label learning.
\newblock {\em Pattern recognition}, 40(7):2038--2048, 2007.

\bibitem{zou2010mining}
Z.~Zou, J.~Li, H.~Gao, and S.~Zhang.
\newblock Mining frequent subgraph patterns from uncertain graph data.
\newblock {\em IEEE Transactions on Knowledge and Data Engineering},
  22(9):1203--1218, 2010.

\end{thebibliography}
